\documentclass[runningheads]{llncs}
\usepackage{multirow}
\usepackage{eccv}

\usepackage{eccvabbrv}

\usepackage{graphicx}
\usepackage{booktabs}

\usepackage[accsupp]{axessibility}  % Improves PDF readability for those with disabilities.

\usepackage{hyperref}

\usepackage{orcidlink}

\newcommand{\esther}[1]{\textcolor{black}{#1}}
\def\ehtim{\texttt{eht-imaging}\xspace}
\def\ngehtsim{\texttt{ngehtsim}\xspace}
\newcommand{\neuralDMD}{\textsc{\small{NeuralDMD}}\xspace}

\DeclareMathOperator*{\argmin}{arg\,min}
\def\ngeht{\texttt{ngEHT}\xspace}
\def\ehtobs{\texttt{EHT2017}\xspace}
\def\ngehtp{\texttt{ngEHT+}\xspace}
\def\sw{\texttt{StarWarps}\xspace}

\newcommand{\varbaseline}{\textsc{\small{3D-Var}}\xspace}
\newcommand{\optdmd}{\textsc{\small{optDMD}}\xspace}
\newcommand{\siren}{\textsc{\small{SIREN}}}
\newcommand{\mfn}{\textsc{\small{MFN}}}
\newcommand{\ffm}{\textsc{\small{FFM}}}

\begin{document}

% \title{NeuralDMD: Interpretable Untrained Neural Network for Dynamic Imaging from Sparse and Noisy Observations} 
\title{NeuralDMD: Interpretable Neural Representation of Dynamics from Sparse and Noisy Measurements} 

\titlerunning{NeuralDMD: Dynamic Imaging from Sparse and Noisy Measurements}

\author{Ali SaraerToosi\inst{1, 2}\orcidlink{0009-0003-4620-8448} \and
Renbo Tu\inst{1, 2}\orcidlink{0009-0002-8835-7225} \and
Esther Y. H. Lin\inst{1, 2}\orcidlink{0000-0002-2260-4345} \and
Kamyar Azizzadenesheli\inst{3}\orcidlink{0000-0001-8507-1868} \and
Aviad Levis\inst{1, 2}\orcidlink{0000-0001-7307-632X}
}

\authorrunning{SaraerToosi.~A. et al.}

\institute{University of Toronto, Toronto, ON, Canada, \email{\{asaraert, tutubo, lin, alevis\}@cs.toronto.edu}
\and
Vector Institute, Toronto, ON, Canada \and
NVIDIA Corporation, USA, \email{kamyar@nvidia.com}\\
}
\maketitle

\begin{figure}[ht]
    \centering
    \includegraphics[width=1.0\linewidth]{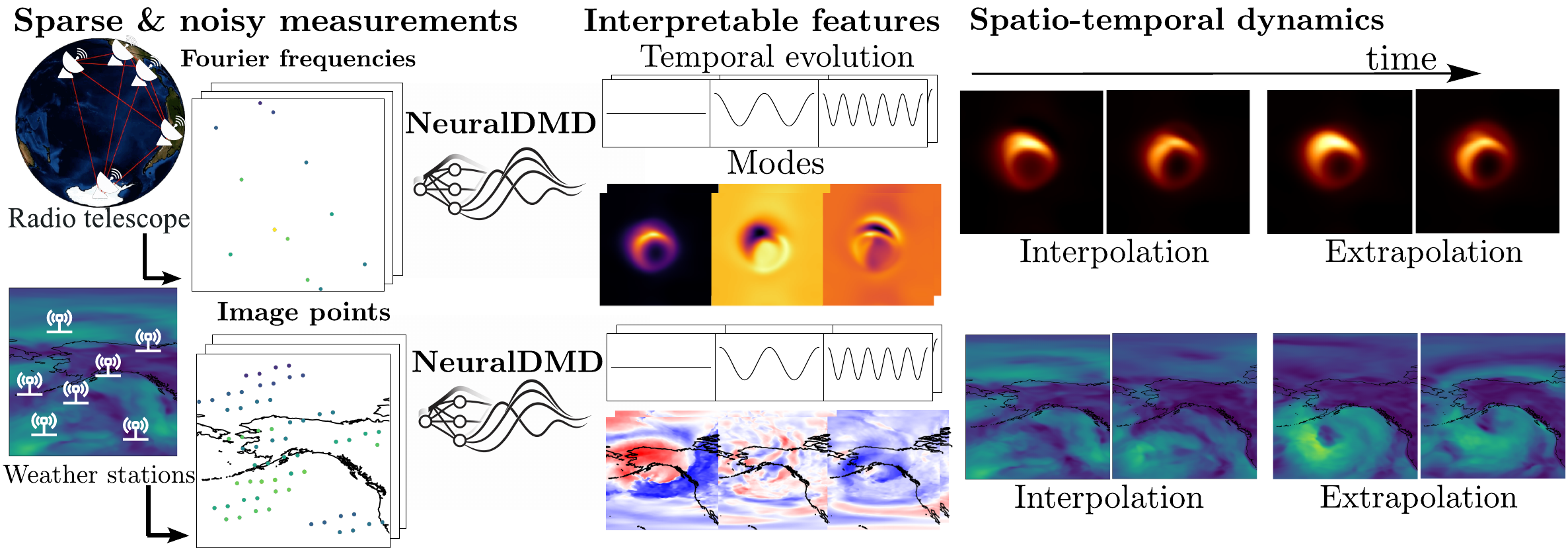}
    \caption{
    \esther{\textbf{\neuralDMD} reconstructs spatio-temporal dynamics from sparse observations in both indirect and direct sensing regimes, including sparse Fourier (top) and image-domain (bottom) measurements. It models the field as a low-rank linear combination of spatial modes with simple spectral time evolution, yielding an interpretable decomposition that supports forecasting. We demonstrate this on black hole imaging (top) and atmospheric data assimilation (bottom).}
    }
    \label{fig: neural dmd}
\end{figure}

\begin{abstract}
Many challenges in scientific imaging involve solving ill-posed inverse problems, where the goal is to recover spatio-temporal fields from indirect, noisy, and highly sparse measurements — often without access to ground truth data or reliable simulators. To address this challenging scenario, we present \neuralDMD, an interpretable, untrained (per-instance) reconstruction framework that combines neural implicit representations with Dynamic Mode Decomposition (DMD) to reconstruct continuous spatio-temporal dynamics directly from measurements. \neuralDMD parameterizes DMD modes as continuous neural fields, and imposes a low-rank linear dynamics prior with spectral time evolution to enforce temporal continuity. This formulation enables both forecasting under sparsity, and yields interpretable modes and spectra.
We find that NeuralDMD outperforms baselines on a wide variety of tasks: from weather data assimilation from sparse station observations to interferometric (Fourier domain) observations of Sagittarius~A*, the black hole at the center of our galaxy. Moreover, \neuralDMD remains stable when extrapolating into the future. While this framework is most naturally suited to linear dynamics, we show that it can be applied to nonlinear regimes, though with extrapolation performance that degrades with increasing nonlinearity.
Together, these results show that \neuralDMD enables interpretable reconstruction and forecasting of spatio-temporal dynamics directly from sparse and indirect measurements without relying on numerical simulators or training data.
\end{abstract}    
\section{Introduction}
\label{sec:intro}
A central challenge in scientific imaging is to recover continuous spatio-temporal dynamics from sparse and noisy measurements.
Such measurements may be direct image-domain samples or indirect observations such as sparse Fourier measurements. 
To further compound the challenge, scientific imaging settings are data-constrained or lack known governing equations, making it difficult to generate simulation data.
In this work, we propose \neuralDMD, a per-instance reconstruction method that supports both direct and indirect sensing regimes. 
We demonstrate it on weather data assimilation (image domain) and black hole imaging from the Event Horizon Telescope (EHT) interferometric observations (Fourier domain).

There have been efforts to address this problem of spatio-temporal reconstruction under sparse and noisy observations without simulation training. Neural fields (coordinate-based Multilayered Perceptrons or MLPs) can model high-dimensional signals of real world physical phenomena \cite{Lin_2025, xie2022neuralfieldsvisualcomputing} but require dense data, lack interpretability, and are sensitive to overfitting under sparse data conditions \cite{Petrini_2023}. To mitigate issues, physics informed neural networks (PINNs) learn explicit analytical equations that can model fluids \cite{neural_implicit_flow}, but assume that the equations are known, making them biased and not suitable when physics is unknown \cite{pinns_failures}. Operator learning approaches are also capable of modeling physical phenomena but suffer from the same simulation-to-real domain gap, making them heavily biased on prior assumptions. Purely data-driven models (no simulation training) such as dynamic mode decomposition (DMD; \cite{dmd_book}) preserve interpretability but are extremely difficult to use under high noise and for sparse data. To the best of our knowledge, no existing approach is simultaneously simulation-free, supports indirect operators, and provides interpretable forecasting prior to extreme sparsity.

We propose \neuralDMD, a model-free approach that enables spatio-temporal reconstruction from sparse, noisy, and indirect measurements. \neuralDMD fits a compact dynamical model directly to observations through the measurement operator. It enforces temporal coherence via a low-rank modal decomposition with explicit form of time evolution, and parameterizes modes as neural fields to enable grid-free evaluation. This yields an interpretable set of modes and spectra and supports forecasting beyond the observation window.

In summary, our contributions include:
\begin{itemize}
    \item We propose a new representation for modeling dynamic physical processes: \neuralDMD.\footnote{Code publicly available here: \href{https://github.com/pi-vision/NeuralDMD.git}{https://github.com/pi-vision/NeuralDMD.git}.}
    \item We show how to optimize this representation on physical data from different imaging scenarios without requiring simulation training.
    \item We thoroughly evaluate the method and show state of the art results.
\end{itemize}
\section{Related Work}
\label{sec: Relevant Prior Work}

We review prior approaches according to their data requirements; emphasizing whether they rely on simulation-trained priors, known governing equations, or densely observed ground-truth states.

\paragraph{\textbf{Simulation-driven modeling.}}
This class of modeling includes operator learning methods \cite{deepONet, TFNO, UFNO, neural_operator,azizzadenesheli2024neural, neural_ode}, principal component analysis (PCA; \cite{primo}), video predictor models \cite{li_video_prediction}, including recent approaches trained on numerical simulations to predict black hole dynamics~\cite{tu2026bhcast}, and some variations of INRs such as CORAL \cite{CORAL_INR} that are designed for operator learning. These approaches, although very expressive for predicting in-distribution dynamics, are not designed for scientific discovery as they are heavily biased by the numerical simulations that were used for training and struggle to express out-of-distribution features.
Moreover, most video predictor models rely on the availability of dense frame sequences and are incapable of reconstructing dynamics from sparse observations.

For \neuralDMD, we explicitly avoid training on simulations and directly fit the sparse and noisy measurements to enable scientific discovery. We only compare \neuralDMD to one simulation-trained model \cite{CORAL_INR} to demonstrate the failure point of such models when the observed physical dynamics differ from what was seen in simulations.
\paragraph{\textbf{Physics-informed modeling}}
This class of models includes physics informed neural networks (PINNs) and some variations of implicit neural representations (INRs) that take advantage of prior physical knowledge of the underlying system.
Existing latent-state approaches mainly (i) estimate parameters of a known model~\cite{Chen2022,physics_inverse_graphics,learning_physics_from_video,material_reconstruction,morpheus}, or (ii) embed predefined governing equations into a network~\cite{neural_implicit_rep,material_reconstruction,morpheus}, which ties dynamics to assumed physics and limits performance under sparse, noisy, or poorly constrained conditions. Similar ideas appear in black hole imaging. BHNeRF~\cite{levis2022gravitationally} and PI-DEF~\cite{Feng_2026_CVPR} recover three-dimensional structure from observations, with PI-DEF additionally recovering the velocity field. In contrast, Levis et al.~\cite{Levis_2021_ICCV} recover image-plane dynamics through a modal decomposition, but assume a known PDE. \neuralDMD instead learns image-plane dynamics directly from observations without assuming a predefined dynamical model.

We do not compare \neuralDMD to this class of models, as they suffer from the same limitations as simulation-driven modeling: they either require simulation data for training or assume access to the PDEs that control the dynamics, a limiting assumption when the goal is scientific discovery rather than parameter estimation under a fixed physical model.

\paragraph{\textbf{Untrained modeling.}}
In this paper, we consider the set of problems within this category. Most INR and DMD variations, including \neuralDMD, are fully data driven, i.e., they rely only on the available data with little to no reliance on numerical simulations. Many scientific problems fall within this category, where it is necessary to minimize physical assumptions for scientific discovery (we discuss some scientific applications in Sec.~\ref{sec: Background}). 

Neural fields have been shown to have immense expressive power\cite{xie2022neural, lin2025learning}. They have been used to reconstruct 3-dimensional videos from a dense or sparse set of camera views \cite{NeRF, D-nerf, sparseNeRF}. However, typically it is assumed that we have complete image-plane measurements, i.e., no missing pixels and minimal measurement noise. In principle, neural fields can be used as a model-free approach to learn dynamics from sparse measurements. However, in practice, direct neural representation of the dynamics is used only when the physics is already known \cite{neural_implicit_rep,material_reconstruction,morpheus}. This is due to the large degree of freedom in a multi-layered perceptron (MLP), where even a small model would have millions of parameters, quickly leading to overfitting. As such, neural fields such as NeRF and dynamic variants~\cite{NeRF,neuralSDF,D-nerf,DynIBaR,4d_gaussian_splatting,gaussian_flow,HexPlane,K_Planes,L4GM}, including sparse-view NeRFs~\cite{sparseNeRF}, still require dense, high-quality observations and typically address only simple physical systems.

Modal decomposition methods offer a complementary perspective by distilling complex dynamics into the evolution of a few characteristic modes, i.e. a small number of degrees of freedom. Dynamic Mode Decomposition (DMD) learns a continuous-time linear operator directly from time-ordered snapshots, constructing a matrix $A$ whose eigen-decomposition yields spatial modes associated with dominant coherent structures~\cite{Schmid2010, Tu2014}. As a result, DMD is model-free and data-driven~\cite{Kutz2016, node_dmd}, and its low-rank structure makes it well-suited for reduced-order modeling and forecasting~\cite{Mezic2013}. Related ideas also arise in dynamical-texture models~\cite{Doretto_2003, layered_DT, variational_layered_DT, DMD_and_DT}, extending texture analysis to stochastic spatio-temporal motion such as smoke, water, and turbulence. However, standard DMD and its variations require access to dense video frames rather than sparse measurements and are sensitive to initialization. Furthermore, because they operate on grids as opposed to continuous fields, their memory and computational costs scale with the number of pixels, making them both memory-intensive and ill-suited for sparse and noisy observations.

\section{Background and Notation}
\label{sec: Background}

We review linear dynamical systems and Dynamic Mode Decomposition, which form the foundation of our approach.

\begin{figure*}[t]
  \centering
  \includegraphics[width=\textwidth]{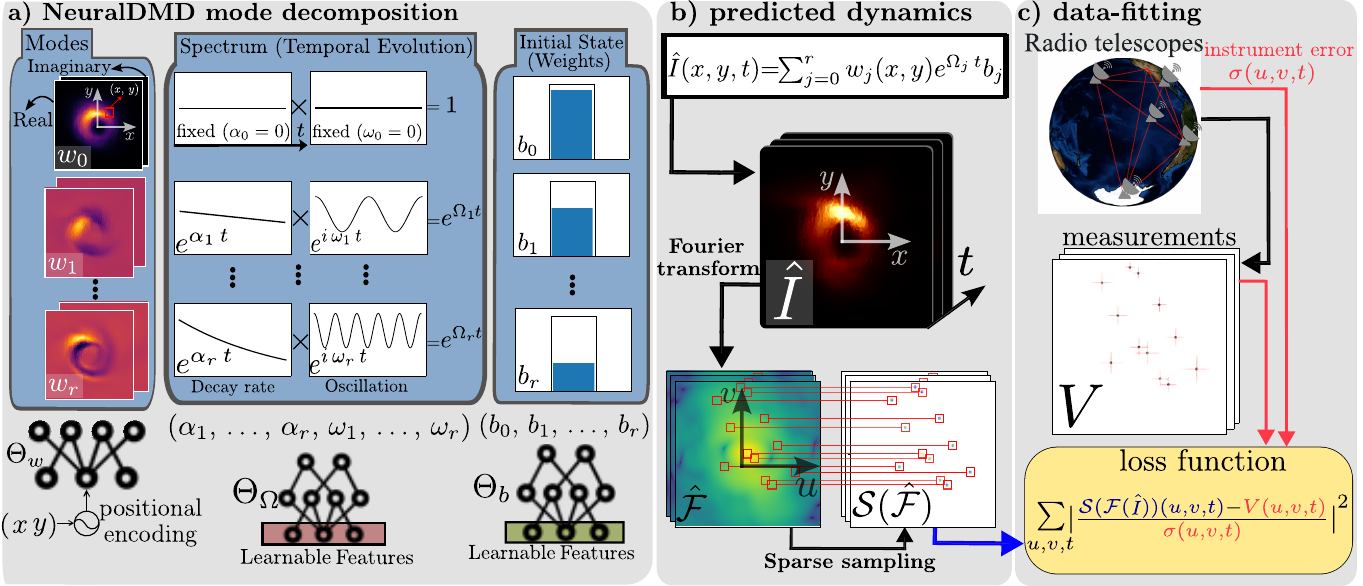}
  \caption{\textbf{(a)} Block diagram of the \neuralDMD architecture. The model comprises three MLPs defining the spatio-temporal dynamics: $\Theta_w$, mapping positionally encoded spatial coordinates $(x,y)$ to mode values $w_j(x,y)$; $\Theta_\Omega$, producing complex spectral components (decay rate $\alpha$, frequency $\omega$) from a learnable latent vector; and $\Theta_b$, mapping a learnable latent vector to the initial state $b_j$ for temporal evolution. \textbf{(b), (c)} The learned modes and coefficients are combined to produce the predicted dynamics $\hat{I}(x, y, t)$. For the black hole setting, predictions are Fourier transformed ($\hat{F}$) and sampled at measured baseline coordinates ($\mathcal{S}(\hat{\mathcal{F}})$). The loss is computed between the observed visibilities $V$ and the corresponding model prediction. For image-domain experiments (weather data assimilation) the loss is instead computed directly on the sparse measurements.}
  \label{fig: forward model}
\end{figure*}

\paragraph{\textbf{Linear dynamical systems.}}
We consider an autonomous linear dynamical system $I_{t+1}=A I_t$, where $A$ governs the temporal evolution of the system state. We assume $A$ to be constant in time (see \cite{Zhang2019} for non-constant A) and no external inputs act on the system, so the dynamics are fully described by the state evolution~\cite{Proctor2016, Proctor2018}. Such systems admit a modal decomposition in which the spatio-temporal field can be expressed as a low-rank superposition of spatial modes evolving in time,
\begin{equation}
I(x,y,t)=\sum_{j=1}^{r}  w_j(x,y)\, a_j(t)  , \quad a_j(t) = b_je^{\Omega_j t} \,
\label{eq:modal}
\end{equation}
where $w_j(x,y)$ are spatial modes, $a_j(t)$ capture the temporal evolution, $b_j$ are coefficients determined by the initial state of the system, and $\Omega_j=\alpha_j+i\omega_j$ are complex spectral coefficients whose real and imaginary parts encode growth/decay rates and oscillation frequencies. Having established the modal form of a linear dynamical system, we now turn to Dynamic Mode Decomposition (DMD), which provides a practical framework for estimating the parameters $\{w_j,\Omega_j\}$ from observation.

\paragraph{\textbf{Dynamic Mode Decomposition (DMD).}
DMD estimates the modal decomposition of a dynamical system directly from time-ordered observations. Classical DMD models the system dynamics as a linear mapping between consecutive states or frames,
\begin{equation}
  [I_1, I_2, \dots, I_M] = A \, [I_0, I_1, \dots, I_{M - 1}] = A \, X,
\end{equation}
where $X$ is a data matrix whose columns are vectorized snapshots of the state at successive time steps. DMD estimates spatial modes $w_j$ and spectral coefficients $\Omega_j$ from the eigendecomposition of a low-rank approximation of $A$, computed using singular value decomposition (SVD)~\cite{original_dmd,dmd_intro,dmd_book}. 
While computationally efficient~\cite{salvador_jasin_2026_19311360}, classical DMD assumes direct access to densely observed system states. In many scientific imaging problems, however, the underlying state is never observed directly and must be reconstructed from indirect measurements.}

\paragraph{\textbf{Optimized DMD (\optdmd).}}
To address these limitations,  \optdmd \cite{optdmd, bag_optdmd} reformulate modal extraction as a nonlinear least-squares problem,
\begin{equation}
  \min_{\mathbf{w}_j,\Omega_j}\;
  \sum_{t\in\{t_1,\dots,t_M\}}
  \bigl\|X_t-\sum_{j=1}^{r} \mathbf{w}_j e^{\Omega_j t} b_j\bigr\|_2^2,
  \label{eq: optdmd}
\end{equation}
where $X_t$ denotes the observed system state at time $t$. The initial-state coefficients are computed as $b_j=\mathbf{w}_j^{\dagger}X_0$, where $\mathbf{w}_j^{\dagger}$ is the complex conjugate transpose of $\mathbf{w}_j$. This formulation improves robustness to noise and directly estimates a continuous-time spectrum $\Omega_j$. However, \optdmd still assumes densely observed states defined on a fixed spatial grid and direct access to $X_0$, limiting its applicability to sparse or indirect measurements.

\section{NeuralDMD}
\label{sec: Methods}

We introduce \neuralDMD, which combines neural representations with the \optdmd formulation by parameterizing the spatial modes $w_j(x,y)$ as neural fields and fitting the model directly to sparse measurements, without requiring densely observed states on a fixed grid, and inferring the initial state from all measurements rather than only those at the initial time.

\subsection{\neuralDMD Representation}
\label{sec:neuraldmd_rep}
Building on the modal representation in Eq.~\eqref{eq:modal}, \neuralDMD extends \optdmd by parameterizing the spatial modes as \emph{continuous} neural fields\cite{NeRF,instantNGP,chang_laplacian, zhaorevealing} with trainable weights $\Theta=[\Theta_w,\Theta_\Omega,\Theta_b]$ and learning the modal parameters directly from sparse measurements. $\Theta_w$, $\Theta_\Omega$, and $\Theta_b$ parameterize the networks for the spatial modes $w_j(x,y)$, spectral coefficients $\Omega_j$, and initial coefficients $b_j$, respectively (Fig.~\ref{fig: forward model}). In particular, we represent each spatial mode as a coordinate multilayer perceptron (MLP) network:
\begin{equation}
w_j(x,y) = w_j(x,y;\Theta_w),
\end{equation}
allowing evaluation at arbitrary spatial coordinates.

To support stable optimization, we treat the spectral parameters and initial coefficients as learnable quantities and parameterize them with lightweight networks:
\begin{equation}
\Omega_j = \Omega_j(\Theta_\Omega), \qquad b_j = b_j(\Theta_b).
\end{equation}
Crucially, in contrast to \optdmd, this formulation eliminates the need for direct access to the initial state $X_0$. Instead $b_j$ is inferred jointly from all measurements, since its evolution under the dynamical model must remain consistent with the entire observation sequence. The resulting prediction $\hat I(x,y,t)$ follows Eq.~\eqref{eq:modal} with the specified neural parameterizations; the next subsections define how $\Theta$ is optimized from sparse (i) direct image-domain observations and (ii) indirect Fourier-domain measurements.

\subsection{Reconstruction from Sparse Image-Domain Observations}
\label{sec:direct_obs}
The problem of reconstruction from direct image domain sparse observations is present in weather data assimilation~\cite{Carrassi2018}. To reconstruct a field from sparse spatio-temporal measurements at coordinates $\mathcal{S}=\{(x_k,y_k,t_k)\}_{k=1}^N$, with corresponding measurements $\{I(x_k,y_k,t_k)\}$, we fit the parameters, $\Theta^{\star}$, by minimizing the residual over observed samples,
\begin{equation}
\Theta^{\star}
=\argmin_{\Theta}\sum_{(x,y,t)\in\mathcal{S}}
\bigl\|I(x,y,t)-\hat I(x,y,t)\bigr\|_2^{2},
\label{eq:neuraldmd_direct_loss}
\end{equation}
where $\hat I(x,y,t)$ is given by Eq.~\eqref{eq:modal} with the neural parameterizations:
\begin{equation}
\hat I(x,y,t) = \sum_{j=1}^{r}  w_j(x,y;\Theta_w) b_j(\Theta_b)e^{\Omega_j(\Theta_\Omega) t} \,
\label{eq:neuraldmd_model}
\end{equation}
To ensure stable dynamics, we constrain the real part of the spectrum to be non-positive, i.e., $\alpha_j=\Re(\Omega_j)\le 0$ (see Supp~\ref{sec:supp_impl}). This enforces physically meaningful decay and oscillatory behavior while preventing exponential growth. After optimization, forecasting can be done by evaluating $\hat I(x,y,t)$ at future times $t$. In Sec.~\ref{sec:experiments}, we explore the accuracy of extrapolation beyond the observed times.

\subsection{Reconstruction from Indirect Fourier Measurements}
In some imaging systems, observations are indirect and obtained as sparse Fourier samples or projections, such as MRI~\cite{MRI_sodickson}, Fourier ptychography~\cite{ptychography_wolfgang}, coded diffraction imaging~\cite{diff_imaging_attal}, computational microscopy~\cite{neural_space_time_model}, and cryo-EM imaging~\cite{CryoDRGN}. In this work, we consider interferometry with the Event Horizon Telescope (EHT), a global interferometer used to image black holes~(see e.g.~\cite{EHT_SgrA_III_2023}).
Each telescope pair measures signal correlations, known as visibilities, which correspond to samples of the underlying source's Fourier transform.
Mathematically, the visibility at spatial frequency $(u,v)$ and time $t$ is given by
\begin{equation}
    V(u,v,t) = \mathcal{F}[I_t](u,v),
    \label{eq:visibility}
\end{equation}
where $I_t$ is the unknown image frame to be reconstructed and $\mathcal{F}$ denotes the 2D Fourier transform.

Observations are available only at a sparse set of Fourier coordinates: $\mathcal{S}_t=\{(u_k,v_k,t_k)\}_{k=1}^N$, which vary over time due to Earth-rotation~\cite{Thompson2017}. We fit \neuralDMD by matching predicted and observed visibilities,
\begin{equation}
\Theta^{\star}
=\argmin_{\Theta}\sum_{(u,v,t)\in\mathcal{S}_t}
\bigl\|V(u,v,t)-\mathcal{F}[\hat I_t](u,v)\bigr\|_2^{2},
\label{eq:neuraldmd_fourier_loss}
\end{equation}
More generally, the Fourier transform in Eq.~\eqref{eq:neuraldmd_fourier_loss} can be replaced by an arbitrary differentiable measurement operator, extending \neuralDMD beyond Fourier imaging to nonlinear observation models and more general inverse problems.

For EHT interferometry, we fit models in the visibility domain using a $\chi^2$ data term, defined as:
\begin{equation}
    \chi^2(\mathbf{\Theta})= \frac{1}{N} \sum_{i = 1}^N \frac{|V_i - \hat{V}_i(\mathbf{\Theta})|^2}{\sigma_i^2},
    \label{eq: r-chi-squared}
\end{equation}
where $N$ is the total number of observations and $\hat{V}_i$ is the model visibility.
In our experiments, fitted models typically achieve $\chi^2 \approx 1$, indicating agreement with the measurements within their uncertainties (see Supp.~\ref{sec:supp_eht_baselines} for details).
\section{Experiments}
\label{sec:experiments}
\subsection{Experimental Details}
\paragraph{\textbf{Experimental Setup}}
\label{subsec:setup}
For all experiments, we fit \neuralDMD by minimizing the appropriate measurement residual (Sec.~\ref{sec: Methods}) 
Here, an \emph{epoch} denotes a fixed number of gradient steps. We uniformly batch over observed coordinate sets (image-domain $\mathcal{S}$ or Fourier-domain $\mathcal{S}_t$), using spatial batch sizes of 64--256 and selecting a per-epoch sampling fraction between 0.1 and 1.0 solely to control memory footprint; results are insensitive to this choice (see Supp). The number of modes $r$ is chosen based on the sparsity (typically $r\in[5,75]$). Ablations on sparsity versus mode count provided in the Supp.

\paragraph{\textbf{Implementation Details}}
\label{subsec:implementation}
We use a fixed architecture for \neuralDMD in all experiments. The mode network $\Theta_w$ comprises four MLP blocks, each consisting of two linear layers with 256 hidden units and SiLU activations. The spectrum and initial-coefficient networks, $\Theta_\Omega$ and $\Theta_b$, share a lightweight architecture: a learnable 16D latent input followed by a two-layer MLP with 64 hidden units (SiLU) and a final linear layer producing per-mode outputs; $\Theta_\Omega$ uses a sigmoid output mapping, while $\Theta_b$ uses a softplus on amplitudes and no activation on phases. Spatial inputs use NeRF-style sinusoidal positional encoding\cite{NeRF} with $L\in[0,10]$ frequencies (powers of two), with smaller $L$ used under higher sparsity. We use Adam optimizer with learning rate $10^{-4}$ and a reduce-on-plateau scheduler for $10^{4}$--$10^{6}$ epochs depending on the experiment. All parameters are randomly initialized except for black hole imaging. There, we use structured initializations (disk pretraining for INR baselines and Zernicke initialization for \neuralDMD), with full details in the Supp.~\ref{sec:supp_eht}.

\paragraph{\textbf{Baselines}}
\label{subsec:baselines}
For image-domain sparse observations (weather), we compare \neuralDMD against a naive data-assimilation baseline: \varbaseline \cite{OI_vs_3DVAR, 3d4dvar}, and two spatio-temporal coordinate networks that regress $I(x,y,t)$ from $(x,y,t)$: a vanilla MLP and SIREN~\cite{siren}. For Fourier-domain sparse observation experiments (black hole imaging), we compare against (i) a spatio-temporal INR baseline fit directly to visibilities, (ii) \optdmd, (iii) StarWarps~\cite{starwarps}, which is a gold-standard dynamical imaging method used in the EHT community~\cite{EHT_SgrA_III_2023}, and (iv) resolve~\cite{jakob_resolve}, a more recent Bayesian image reconstruction method.

\paragraph{\textbf{Ablations and Evaluation Metrics}}
\label{subsec:eval_metrics}
We ablate the INR backbone (SIREN \cite{siren}, MFN\cite{mfn}, Fourier-feature MLP\cite{ffm}) and the parameterization of $(\Omega,b)$, direct optimization versus networks $\Theta_\Omega,\Theta_b$. For fairness, all INR baselines use the same MLP backbone as \neuralDMD’s mode network $\Theta_w$, with time as an additional input and comparable positional encoding. All methods are fit under the same observation operator and sampling pattern. We report RMSE, PSNR, SSIM, and LPIPS for all reconstructions.

\subsection{Image-Domain Sparse Observations}
\label{subsec:image_domain}

\paragraph{\textbf{Weather Data Assimilation.}}
\label{par:weather}

\begin{table}[t]
\centering
\footnotesize
\setlength{\tabcolsep}{4pt}
\renewcommand{\arraystretch}{1.05}
\caption{\textbf{Quantitative evaluation: Weather data assimilation (ERA5).} Quantitative comparison (mean$\pm$std over frames) for dense ($\sim 5000$) and sparse ($\sim 500$) observations per frame. We report results at both sparsity levels to highlight performance in a highly undersampled, operationally realistic ($\sim 500$) and in a denser regime ($\sim 5000$), where \varbaseline is applicable. \neuralDMD achieves the best overall performance at both sparsity levels.}
\label{tab:weather_quant}
\resizebox{0.9\linewidth}{!}{
\begin{tabular}{llcccc}
\toprule
Method & Samples per frame & RMSE$\downarrow$ & PSNR$\uparrow$ & SSIM$\uparrow$ & LPIPS$\downarrow$ \\
\midrule

\multirow{2}{*}{\textbf{\neuralDMD (ours)}}
& $\sim 500$  & \textbf{2.3$\pm$0.1} & \textbf{19.1$\pm$0.8} & \textbf{0.53$\pm$0.03} & \textbf{0.59$\pm$0.02} \\
& $\sim 5000$ & \textbf{1.5$\pm$0.1} & \textbf{24.7$\pm$0.5} & \textbf{0.70$\pm$0.01} & \textbf{0.32$\pm$0.01} \\
\midrule

\multirow{2}{*}{\varbaseline}
& $\sim 500$  & - & - & - & - \\
& $\sim 5000$ & 3.6$\pm$0.8 & 18$\pm$4 & 0.48$\pm$0.09 & 0.5$\pm$0.1 \\
\midrule

\multirow{2}{*}{Vanilla MLP}
& $\sim 500$  & 2.7$\pm$0.2 & 17$\pm$1 & 0.49$\pm$0.04 & 0.65$\pm$0.03 \\
& $\sim 5000$ & 2.3$\pm$0.2 & 21.0$\pm$0.6 & 0.56$\pm$0.01 & 0.6$\pm$0.2 \\
\midrule

\multirow{2}{*}{\siren}
& $\sim 500$  & 3.0$\pm$0.3 & 17$\pm$1 & 0.50$\pm$0.04 & 0.63$\pm$0.02 \\
& $\sim 5000$ & 2.3$\pm$0.3 & 21$\pm$1 & 0.63$\pm$0.02 & 0.52$\pm$0.02 \\
\bottomrule
\end{tabular}
}
\end{table}

\begin{figure}[t]
\centering
\includegraphics[width=1.0\linewidth]{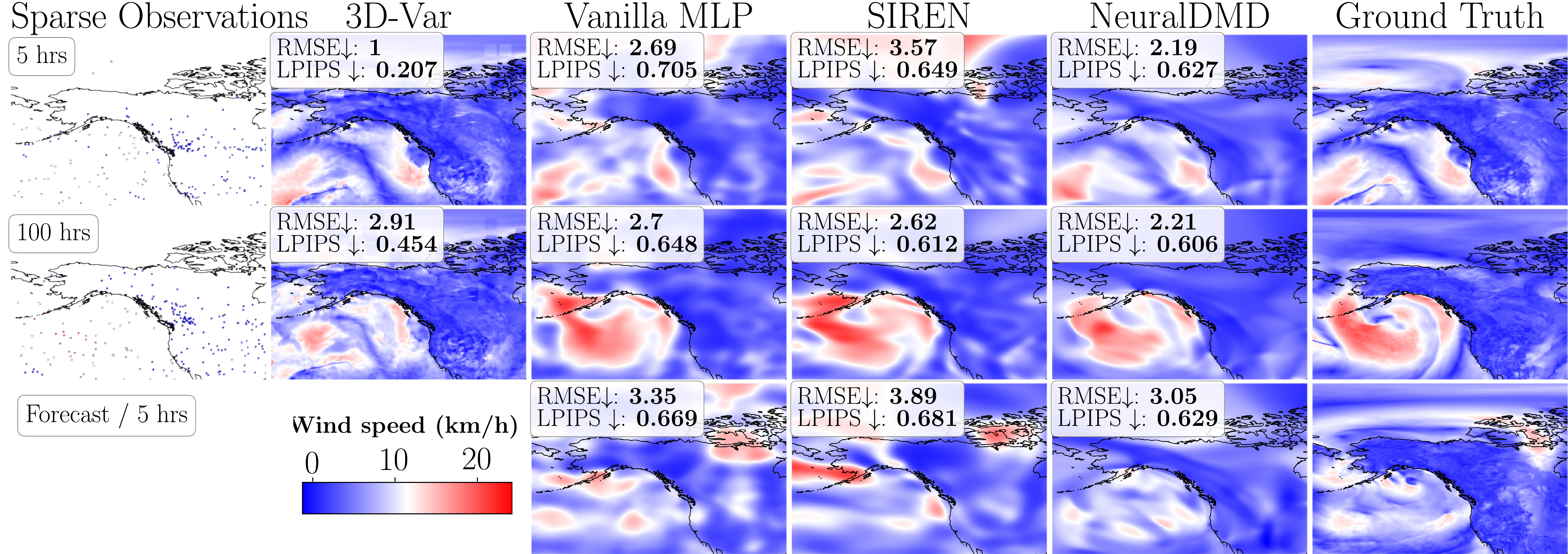}
\caption{\textbf{Qualitative evaluation: Weather data assimilation (ERA5).} Rows show $t{=}5$ hrs, $t{=}100$ hrs, and extrapolation, with $\sim 500$ samples per-frame. \varbaseline is accurate near initialization but degrades over time and cannot extrapolate. \neuralDMD better preserves storm structure and forecasts more reliably than INR baselines, which lack explicit dynamics.}
\label{fig:weather_qual}
\end{figure}

We evaluate \neuralDMD on atmospheric data assimilation, reconstructing a complete spatio-temporal wind field $I(x,y,t)$ from image-domain observations at sparse station locations. We use hourly ERA5 wind-speed magnitude at $10$m above the surface over North America from April~1--7,~2025 (168 frames; \cite{WMO_OSCAR_Surface}), and sample observations at $\sim 5000$ real station coordinates per frame derived from WMO OSCAR/WIGOS \cite{WMO_WIGOS_Guide_2019, WMO_OSCAR_Surface}; with some stations dropping in and out on different days as their operational status changes. We fit models on the assimilation window and evaluate both reconstruction within the window and nowcasting beyond it (April~8,~2025). We compare against a \varbaseline baseline and spatio-temporal INR baselines (vanilla MLP and SIREN~\cite{siren}) fit to the same sparse samples.

We report image metrics computed against dense ERA5 ground truth (mean $\pm$ std across frames). As shown in Fig.~\ref{fig:weather_qual} and Tab.~\ref{tab:weather_quant}, \neuralDMD achieves the lowest error across all metrics and maintains higher-quality reconstructions under sparsity, while also providing stable short-term forecasts beyond the observation window (additional sparsity studies are provided in the Supp.).

\paragraph{\textbf{Wave Equation: Simulation-Trained vs.\ Untrained Modeling.}}
\label{par:wave}
Operator-learning models leverage large numerical simulation datasets to learn reusable solution operators, but their performance can degrade under distribution shift. Here we compare this paradigm against per-instance fitting with \neuralDMD in a sparse-observation inverse problem.

We generate 2D wave-equation sequences and compare \neuralDMD against CORAL~\cite{CORAL_INR}, a representative operator-learning baseline trained on 5000 64-frame videos of low-frequency waves of resolution $96\times 96$. At test time, we evaluate both an in-distribution (ID) regime within the training frequency range and an out-of-distribution (OOD) regime containing higher spatial frequencies, while keeping the sparse sampling pattern fixed across methods. Table~\ref{tab: 2dwave equation} shows quantitative results (qualitative results in Supp.), illustrating that CORAL generalizes well in-distribution (PSNR = 39) but can degrade under frequency shift, whereas \neuralDMD remains reliable because it is fit directly to the measurements of each sequence (details in the Supp.~\ref{sec:supp_wave_coral}). This highlights a key advantage of per-instance fitting, particularly relevant in scientific discovery settings, where the true dynamics may be only partially understood and representative training simulations are often unavailable.
\begin{table}[t]
\footnotesize
\centering
\caption{In distribution (ID) and out of distribution (OOD) quantitative performance of CORAL vs. \neuralDMD. Sparsity rates of $10\%$ and $40\%$ of the total image pixels are used for fitting for ID and OOD cases respectively.}
\label{tab: 2dwave equation}
\setlength{\tabcolsep}{3pt}
\resizebox{0.85\linewidth}{!}{
\begin{tabular}{lcccc}
\toprule
 & CORAL ID & \neuralDMD ID & CORAL OOD & \neuralDMD OOD \\
\midrule
RMSE $\downarrow$
& 0.02$\pm$0.01
& \textbf{(3.6$\pm$0.8)e-3 }
& 0.05$\pm$0.02 
& \textbf{0.014$\pm$0.004 }
 \\

PSNR $\uparrow$
& 39$\pm$1 
& \textbf{56$\pm$1.3 }
& 30$\pm$0.3 
& \textbf{43$\pm$1} \\

SSIM $\uparrow$
& 0.89$\pm$0.0  
& \textbf{0.9994$\pm$0.0002} 
& 0.69$\pm$0.05 
& \textbf{0.997$\pm$0.001} \\

LPIPS $\downarrow$
& 0.012$\pm$0.001
& \textbf{0.0002$\pm$0.0001}
& 0.020$\pm$0.001 
& \textbf{0.002$\pm$0.001 }
\\
\bottomrule
\end{tabular}
}
\end{table}
\subsection{Fourier-Domain Sparse Observations}
\label{subsec:fourier_domain}
We evaluate \neuralDMD on black hole video reconstruction from sparse interferometric measurements. We simulate EHT observations using the array configuration and observing schedule from the 2017 Sagittarius A* campaign, and report reconstruction and extrapolation performance against established baselines.

\paragraph{\textbf{Interferometric Observation Model Simulations.}}
\label{par:grmhd_setup}
We use physically realistic General Relativistic Magnetohydrodynamics (GRMHD) simulations of black hole accretion flows~\cite{ipol_paper,H-ARM_Gammie_2003,grmhd_survey} for evaluation. Corresponding ray-traced image sequences ($I_t(x,y)$) are used as ground truth and passed through an interstellar scattering screen to mimic the Galactic-center observing conditions~\cite{Johnson_2018,Issaoun_2021}.

To simulate sparse visibilities, we use \ehtim~\cite{eht-imaging} together with \ngehtsim \cite{Pesce_ngehtsim_2023} which model Earth-rotation synthesis and sparse Fourier sampling of the EHT 2017 array configuration for Sgr~A$^\ast$ (\ehtobs;~\cite{sgrA_paper1}). This yields a set of complex visibilities $\{V_i\}$ at time-varying spatial frequencies $\{(u_i,v_i,t_i)\}$ with associated uncertainty estimates $\{\sigma_i\}$. Full simulation details, uv-coverage patterns, and additional configurations (e.g., edge-on views and denser arrays) are provided in Supp.~\ref{sec:supp_eht}.

\begin{figure}[t]
    \centering
    \begin{subfigure}{0.55\linewidth}
        \centering
        \includegraphics[width=\linewidth]{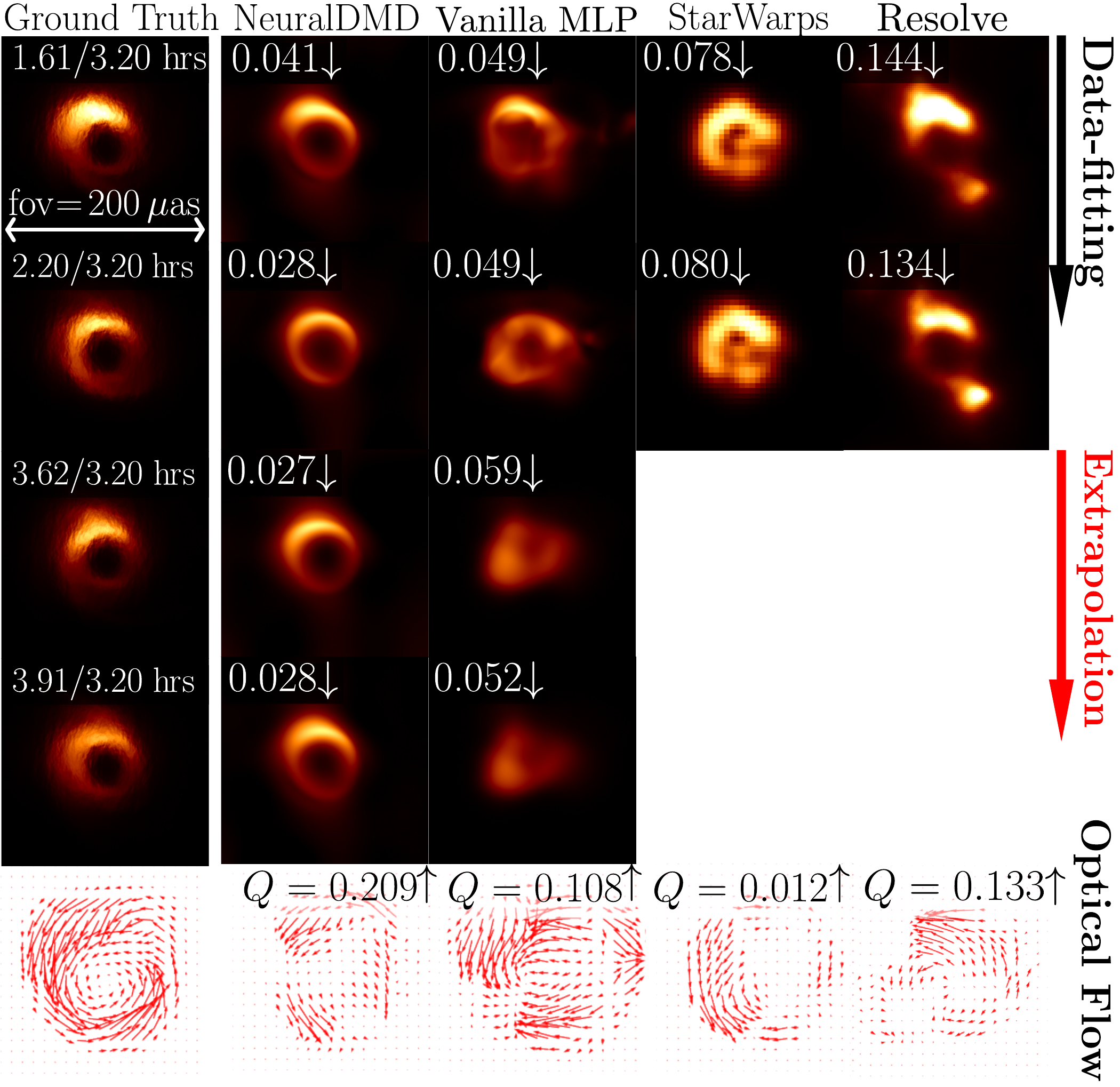}
        \caption{GRMHD reconstruction from sparse interferometric data.}
        \label{fig: GRMHD2 tests EHT2017}
    \end{subfigure}
    \hfill
    \begin{subfigure}{0.42\linewidth}
        \centering
        
        \begin{subfigure}{\linewidth}
            \centering
            \includegraphics[width=\linewidth]{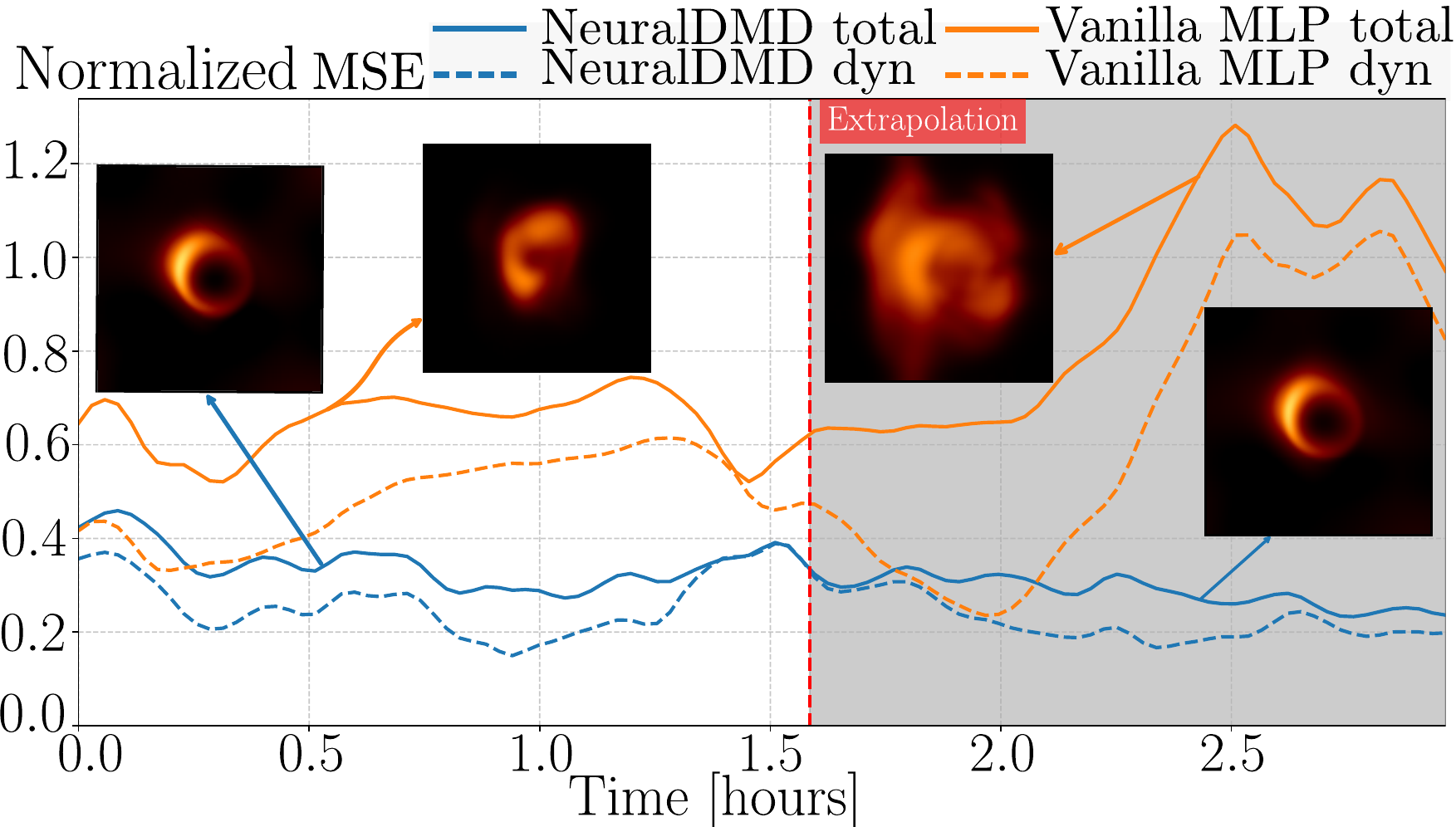}
            \caption{Reconstruction error over time.}
            \label{fig: extra_err EHT2017}
        \end{subfigure}
        
        \vspace{4pt}
        
        \begin{subfigure}{\linewidth}
            \centering
            \includegraphics[width=\linewidth]{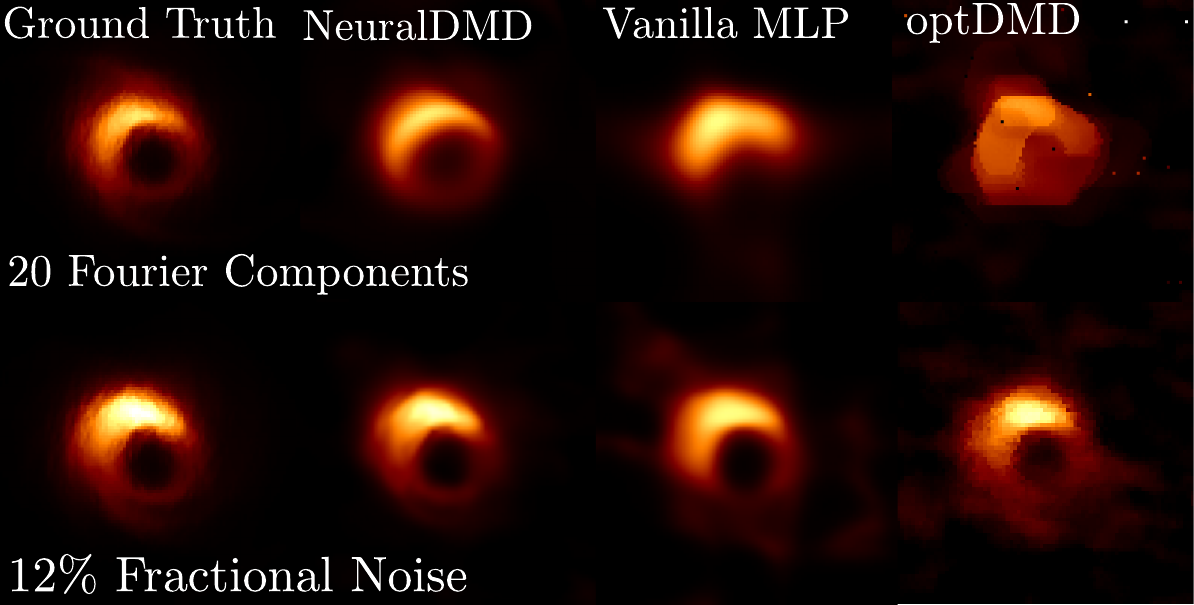}
            \caption{Robustness to noise and sparsity.}
            \label{fig: noise_sparsity_study}
        \end{subfigure}
        
    \end{subfigure}

    \caption{\textbf{Spatio-temporal reconstruction from sparse EHT 2017 visibilities.}
    \textbf{(a)} Reconstructions of a GRMHD sequence from sparse observations during fitting and extrapolation (forecasting). Values denote image-space MSE (lower is better). \neuralDMD produces lower error and better temporal consistency than \sw, resolve, and the spatio-temporal INR (vanilla MLP); The bottom row shows the average optical flow, with $Q$ denoting cosine similarity to the ground truth flow (higher is better).
    \textbf{(b)} Per-frame normalized image-space MSE. Gray region denotes the extrapolation window, where \neuralDMD remains stable in contrast to the vanilla MLP diverges.
    \textbf{(c)} Robustness under increased visibility noise and reduced Fourier sampling. Fractional noise is relative to the data amplitude. A $100 \times 100$ Fourier grid is used; $20$ Fourier components correspond $0.2\%$ sparsity.}
    \label{fig: combined_eht2017_all}
\end{figure}

\paragraph{\textbf{Reconstruction Results and Comparisons.}}
\label{par:eht_reconstructions} Figure~\ref{fig: combined_eht2017_all} compares \neuralDMD against StarWarps~\cite{starwarps}, resolve~\cite{jakob_resolve}, \optdmd, and a spatio-temporal INR baseline (vanilla MLP). Under the extremely sparse \ehtobs sampling, \neuralDMD yields lower reconstruction error and sharper, more temporally consistent reconstructions throughout the data-fitting window. This suggests that representing the dynamics through a small number of modes provides a strong inductive bias for reconstructing dynamics. Additional baseline sweeps and robustness studies under varying noise, sparsity, and corruptions are provided in Supp.~\ref{sec:supp_eht_ablations}. To ensure a fair comparison, we use random initialization with the same seed values for \neuralDMD and the vanilla MLP, and use the same architecture for all baselines (see Supp. for details). In real-world setting, each station observes through a different, time-varying atmospheric column and instrumental chain, introducing unknown complex-valued gain corruptions (affecting both amplitude and phase) in the measured signal. To handle these realistic station-based effects, we marginalize over the complex gains during fitting (details in Supp.~\ref{sec:supp_eht} and \ref{supp:gain_marginalization}).

\paragraph{\textbf{Forecasting Beyond Observations.}}
\label{par:eht_forecast}
After fitting the model to the observations, we forecast future frames by evaluating the learned dynamics at times beyond the last measurement. In the EHT 2017 setting, we fit the model on the first 1.6 hours forecast the subsequent 1.4 hours. Figure~\ref{fig: extra_err EHT2017} shows the per-frame normalized image-space $L_2$ error $\|\hat I_t-I_t\|_2^2/\|I_t\|_2^2$, with the extrapolation region shaded in gray. \neuralDMD continues to produce coherent motion throughout the forecast horizon, whereas the vanilla MLP rapidly diverges once observations are no longer available. Since StarWarps estimates frames only within observation interval, it does not support forecasting.

\paragraph{\textbf{Interpretability.}}
\label{par:eht_interpretability}
Beyond reconstruction quality, \neuralDMD recovers spatial modes $\{w_j(x,y)\}$ together with a continuous-time spectrum $\{\Omega_j=\alpha_j+i\omega_j\}$. The modes capture dominant spatial structures, while the spectrum characterizes their temporal evolution through decay rates and oscillation frequencies. In the black hole setting, for example, the learned frequencies provide direct estimates of characteristic dynamical timescales, such as the angular velocity of rotating structures (examples provided in the Supp. ~\ref{sec:supp_eht}).

\subsection{Robustness and Ablations}
\label{subsec:robustness_ablations}
Table~\ref{tab:eht_ablation_compact} ablates coordinate-network backbones and positional-encoding bandwidth on EHT 2017 reconstructions. Here, $L$ denotes the number of sinusoidal positional-encoding frequencies used for coordinate inputs; we keep the number of layers and channels fixed across all INRs. For each method we report a low- and high-bandwidth setting ($L_\text{low}\,|\,L_\text{high}$), and results across a wider range of bandwidths can be found in the Supp.~\ref{sec:supp_eht_ablations}. Reported values are mean$\pm$std over 5 random initializations and measure reconstruction quality on recovered image frames (not forecasting). Across all backbone and bandwidth settings, \neuralDMD achieves the best overall reconstruction quality, indicating that the dynamical inductive bias provides benefits beyond the choice of INR architecture alone. We also ablate the parameterization of $(\Omega,b)$ by directly optimizing $\{\Omega_j,b_j\}$ as learnable parameters rather than neural network outputs, while keeping the same mode network and measurement loss (``Direct $\Omega/b$''). Directly optimizing $\{\Omega_j,b_j\}$ yields the weakest performance across all metrics, highlighting that over-parameterizing these quantities as neural network outputs is crucial for effective optimization. We further evaluate the robustness of \neuralDMD under varying noise levels and observation sparsity, demonstrating strong reconstruction performance in challenging measurement regimes (Supp.~\ref{sec:supp_eht}).

\begin{table}[t]
\centering
\footnotesize
\setlength{\tabcolsep}{3pt}
\renewcommand{\arraystretch}{1.05}
\caption{\textbf{EHT 2017 ablations.} Reconstruction metrics for different coordinate-network backbones, positional-encoding bandwidths $L$, and the direct optimization baseline for $(\Omega,b)$.}
\label{tab:eht_ablation_compact}
\resizebox{\linewidth}{!}{%
\begin{tabular}{llcccc}
\toprule
Method & $L_\text{low}\,|\,L_\text{high}$ & RMSE$\downarrow$ & PSNR$\uparrow$ & SSIM$\uparrow$ & LPIPS$\downarrow$ \\
\midrule
\multirow{2}{*}{\siren} 
& 3  & 0.072$\pm$0.011  & 23.0$\pm$1.2  & 0.58$\pm$0.032 & 0.29$\pm$0.026 \\
& 10 & 0.075$\pm$0.0058 & 23.0$\pm$0.68 & 0.56$\pm$0.019 & 0.32$\pm$0.011 \\
\midrule
\multirow{2}{*}{\mfn} 
& 3  & 0.13$\pm$0.0095 & 18.0$\pm$0.68 & 0.54$\pm$0.0087 & 0.30$\pm$0.0036 \\
& 10 & 0.15$\pm$0.025  & 17.0$\pm$1.6  & 0.53$\pm$0.020  & 0.32$\pm$0.022 \\
\midrule
\multirow{2}{*}{\ffm} 
& 256  & 0.093$\pm$0.0037 & 21.0$\pm$0.33 & 0.42$\pm$0.0061 & 0.66$\pm$0.018 \\
& 1024 & 0.094$\pm$0.0047 & 21.0$\pm$0.42 & 0.47$\pm$0.031  & 0.59$\pm$0.017 \\
\midrule
\multirow{2}{*}{Vanilla MLP} 
& 3  & 0.051$\pm$0.0029 & 26.0$\pm$0.55 & 0.81$\pm$0.013 & 0.22$\pm$0.0040 \\
& 10 & 0.17$\pm$0.12    & 17.0$\pm$6.0  & 0.63$\pm$0.084 & 0.28$\pm$0.042 \\
\midrule
\multirow{2}{*}{Direct $\Omega/b$} 
& 2  & 0.13$\pm$0.0039 & 18.0$\pm$0.26 & 0.30$\pm$0.011   & 0.41$\pm$0.029 \\
& 10 & 0.19$\pm$0.0043 & 14.0$\pm$0.20 & 0.18$\pm$0.0002  & 0.63$\pm$0.013 \\
\midrule
\multirow{2}{*}{\textbf{\neuralDMD (ours)}} 
& \textbf{2}  & \textbf{0.0349$\pm$0.010} & \textbf{29.18$\pm$1.1} & \textbf{0.6557$\pm$0.099} & \textbf{0.1905$\pm$0.0803} \\
& \textbf{10} & \textbf{0.066$\pm$0.0033} & \textbf{24.0$\pm$0.47} & \textbf{0.54$\pm$0.0021}  & \textbf{0.31$\pm$0.040} \\
\bottomrule
\end{tabular}
}
\end{table}

\begin{figure}[t]
    \centering
    \includegraphics[width=0.75\linewidth]{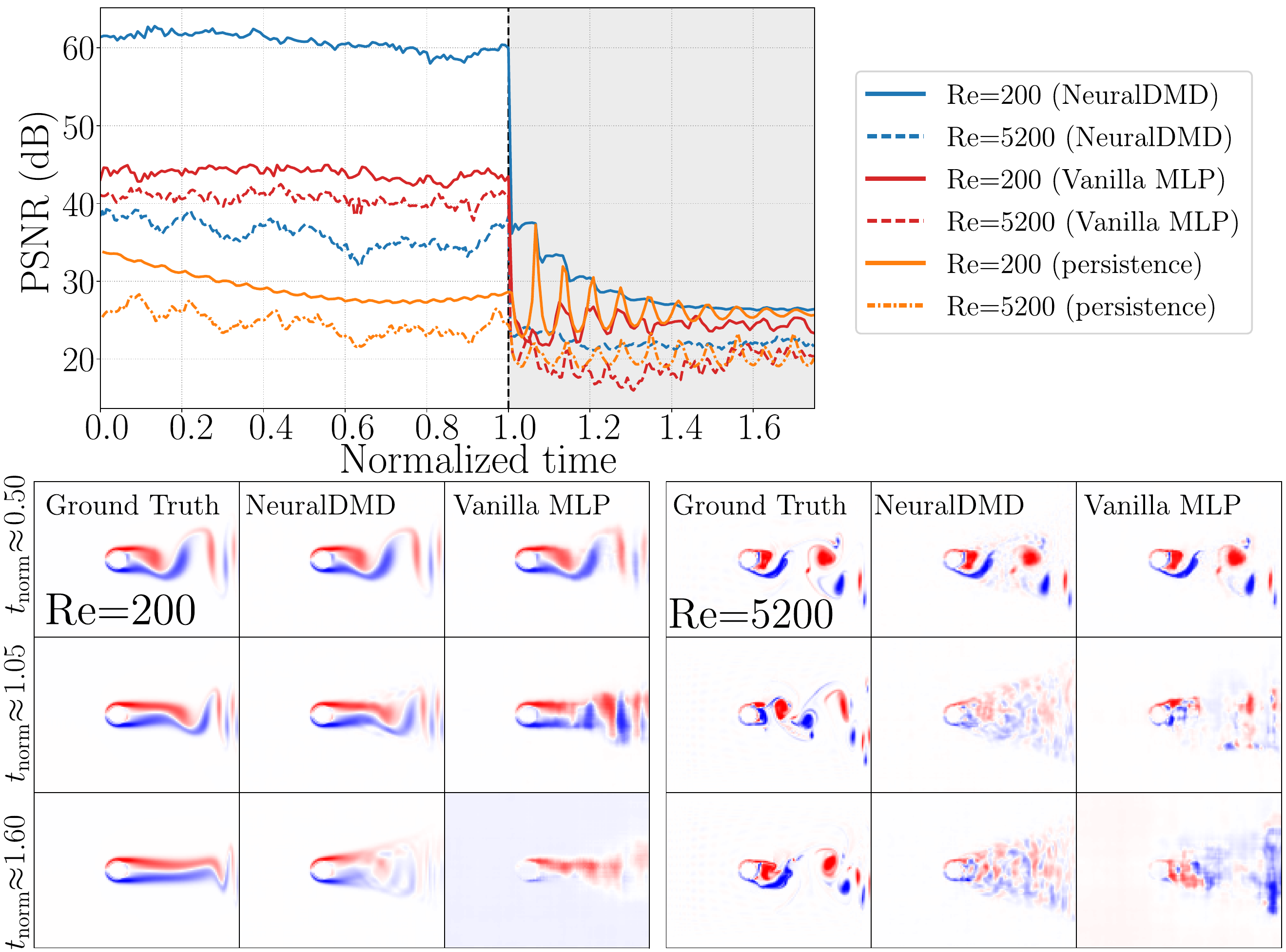}
    \caption{\textbf{top:} PSNR vs. normalized time for Navier-Stokes cylinder flow simulations at Re$=5200$ and Re$=200$. The white and gray regions denotes the observation and forecasting regions respectively. \textbf{bottom:} Ground truth and \neuralDMD reconstructions at $t_{\text{norm}} = 0.5,~1.05,~1.60$. Error grows more rapidly for the highly non-linear case of Re$=5200$ than Re$=200$ (the same fact is reflected in the PSNR plot).}
    \label{fig: cylinder flow}
\end{figure}
\subsection{Limitations: Breakdown of the Linear Dynamics Assumption}
\label{subsec:limitations_cylinder}
A key assumption of \neuralDMD and other DMD-based methods is that the observed dynamics are sufficiently linear over the observation window~\cite{schmid2010dmd,Tu2014}. We study the limits of this assumption using Navier--Stokes cylinder-flow simulations~\cite{schafer1996benchmark}, generated with the \texttt{jax-fluids} library~\cite{bezgin2022jaxfluids}, whose  nonlinearity increases with Reynolds number. The PDE for this systems is
\begin{equation}
    \frac{\partial \mathbf{u}}{\partial t} + (\mathbf{u} \cdot \nabla)\mathbf{u} = - \nabla p + \frac{1}{\text{Re}} \nabla^2 \mathbf{u},\,\, \nabla \cdot \mathbf{u} = 0,
\end{equation}
where $\mathbf{u}$ is the fluid velocity field, $p$ is the pressure, and Re is the Reynolds number. This system is nonlinear due to the convective term $(\mathbf{u}\cdot\nabla)\mathbf{u}$. As Re increases the flow transitions from smooth laminar behavior to increasingly turbulent dynamics, providing a controlled way to study the breakdown of low-dimensional linear representations.

We evaluate \neuralDMD on two cylinder-flow regimes at Re$=200$ and Re$=5200$, which exhibit increasing levels of nonlinear dynamical complexity. For reconstructions, we use 120 and 150 modes, respectively. Figure~\ref{fig: cylinder flow} reports PSNR versus normalized time, where $t_{\mathrm{norm}}\in[0,1]$ is the data-fitting window and $t_{\mathrm{norm}}>1$ is extrapolation. We compare against a \emph{persistence}~\cite{persistence} baseline, which uses the previous frame during the observation window ($\hat I_{t}=I_{t-1}$), and the final observed frame for extrapolation. We define the failure point as when \neuralDMD matches the persistence PSNR.

While \neuralDMD achieves high fidelity during fitting (PSNR $\sim60$ for Re$=200$ and $\sim40$ for Re$=5200$), its forecasting accuracy deteriorates as nonlinearity increases. For both Reynolds numbers, \neuralDMD becomes comparable to the persistence baseline at longer horizons, around $t_{\mathrm{norm}}\approx1.6$, indicating the eventual breakdown of the low-dimensional linear approximation. At Re$=5200$, a vanilla MLP can attain higher PSNR during fitting, but with less reliable extrapolation, consistent with its lack of an explicit dynamical prior.\textbf{}
\section{Discussion}
\label{sec: Discussion}
We introduce \neuralDMD, a simulation-free and interpretable framework for recovering spatio-temporal dynamics from sparse, noisy, and indirect observations while enabling forecasting beyond the observation window. Although we demonstrated the method using image domain and sparse Fourier measurements, the framework naturally extends to arbitrary differentiable observation operators, including dynamic acquisition processes whose measurements evolve on the same timescale as the underlying scene. By combining neural representations with the inductive bias of dynamical systems, NeuralDMD consistently outperforms conventional neural representations. A key feature of NeuralDMD is its explicit low-rank dynamical structure. Although the modeled dynamics are linear, many complex systems admit accurate low-dimensional approximations over finite time windows, yielding interpretable modes and spectra that capture their dominant spatio-temporal behavior.

More broadly, we believe \neuralDMD is particularly relevant for scientific imaging problems where the governing physics remains uncertain and faithful simulations are unavailable. In these settings, learning dynamics directly from observations allows the data to inform, refine, or challenge existing physical models rather than constraining inference to the behavior pre-encoded in simulations. We view this as an important step toward imaging algorithms that not only reconstruct phenomena but also help reveal the underlying physics and enable scientific discovery.
\paragraph{Acknowledgements.} We thank S. Nousias, A. Broderick, M. Shalaby, A. Fuentes, M. Foschi, S. Rashidi, R. Dahale, D. Posselt, and M. Als for helpful comments, and E. Schnetter and D. Lang for their provision of computational resources. A. Levis’s work is supported by the Natural Sciences and Engineering Research Council of Canada (NSERC). This work was also supported by the Ontario Research Fund – Research Excellence under Project Number RE012-045.

\clearpage
\setcounter{section}{0}
\renewcommand{\thesection}{S\arabic{section}}
\setcounter{figure}{0}
\renewcommand{\thefigure}{S\arabic{figure}}
\setcounter{table}{0}
\renewcommand{\thetable}{S\arabic{table}}
\title{Supplementary Material\\
NeuralDMD: Interpretable Untrained Neural Network for Imaging from Sparse and Noisy Observations}
\titlerunning{NeuralDMD: Imaging from Sparse and Noisy Observations}

\author{Ali SaraerToosi\inst{1, 2}\orcidlink{0009-0003-4620-8448} \and
Renbo Tu\inst{1, 2}\orcidlink{0009-0002-8835-7225} \and
Esther Lin\inst{1, 2}\orcidlink{0000-0002-2260-4345} \and
Kamyar Azizzadenesheli\inst{3}\orcidlink{0000-0001-8507-1868} \and
Aviad Levis\inst{1, 2}\orcidlink{0000-0001-7307-632X}
}

\authorrunning{SaraerToosi.~A. et al.}
\institute{University of Toronto, Toronto, ON, Canada \email{\{asaraert, tutubo, lin, alevis\}@cs.toronto.edu}
\and
Vector Institute, Toronto, ON, Canada \and
NVIDIA Corporation, USA \email{kamyar@nvidia.com}\\
}

\setcounter{page}{1}
 \label{sec: supp}
 \maketitle

\section{Implementation Details}
\label{sec:supp_impl}

\paragraph{\textbf{Network parameterization.}
\neuralDMD uses three lightweight MLPs (Fig.~\ref{fig: forward model}).
The \emph{modal network} parameterized by $\Theta_w$ maps spatial coordinates $(x,y)$ to the set of spatial modes $\{w_j(x,y)\}_{j=0}^{r}$.
The \emph{spectral network} parameterized by $\Theta_\Omega$ outputs the complex spectrum $\{\Omega_j\}_{j=1}^{r}$, where $\Omega_j=\alpha_j+i\omega_j$.
The \emph{initial-state network} parameterized by $\Theta_b$ outputs the coefficients $\{b_j\}_{j=0}^{r}$.
All parameters are jointly optimized, with $\Theta=[\Theta_w,\Theta_\Omega,\Theta_b]$.}

\paragraph{\textbf{Positional encoding.}
To represent $w_j(x,y)$ as a continuous neural field, the modal network takes as input a positional encoding of $(x,y)$.
We use a low encoding degree ($L=2$) to regularize recovery of smoothly varying fields under sparse observations (details of the encoding follow standard Fourier-feature mappings).}

\paragraph{\textbf{Stable spectrum parameterization.}
To prevent unstable (explosively growing) dynamics, we constrain the decay rates to be non-positive.
Concretely, we parameterize the real and imaginary parts of the spectrum using bounded activations:
\begin{equation}
\alpha_j \in [-200,\,0], \qquad \omega_j \in [0,\,200],
\end{equation}
implemented via a scaled sigmoid (or equivalent bounded mapping) applied to unconstrained network outputs.
This enforces physically plausible decay and oscillatory behavior while avoiding exponential growth.}

\paragraph{\textbf{Real-valued signals and conjugate pairs.}
Because the underlying spatio-temporal field is real-valued, complex modes appear in conjugate pairs.
We therefore form the reconstruction by taking the real part of the complex modal sum:
\begin{equation}
\hat{I}(x,y,t)= w_0(x,y)\,b_0 \;+\; 2\,\Re\!\left(\sum_{j=1}^{r} w_j(x,y)\,e^{\Omega_j t}\,b_j\right).
\label{eq:supp_real_recon}
\end{equation}
The $j=0$ term is treated as a (real) static component that captures the time-averaged signal; we set $\Omega_0=0$.}

\paragraph{\textbf{Optimization.}
We optimize $\Theta$ by minimizing the measurement residual over the observed sample set (image-domain or Fourier-domain, depending on the measurement operator), using standard stochastic gradient methods.
In all experiments, the loss is evaluated only at observed coordinates (e.g., $(x,y,t)\in\mathcal{S}$ in the image domain, or $(u,v,t)\in\mathcal{S}_t$ in the Fourier domain).}

\section{Mathematics of Classical Dynamic Mode Decomposition}
This section provides the necessary background for modeling the evolution as a linear dynamical system as an interpretable evolution of combination of a set of modes. In classical DMD, given a sequence of snapshots of a dynamical system,
\begin{equation}
  \mathbf{I}_1,\, \mathbf{I}_2,\, \dots,\, \mathbf{I}_m \quad \text{with } \mathbf{I}_k \in \mathbb{R}^n,
\end{equation}
we construct two data matrices
\begin{align}
  X &= \begin{bmatrix}\mathbf{I}_1 & \mathbf{I}_2 & \cdots & \mathbf{I}_m\end{bmatrix}, \\
  X' &= \begin{bmatrix}\mathbf{I}_2 & \mathbf{I}_3 & \cdots & \mathbf{I}_{m+1}\end{bmatrix}.
  \label{eq: X and X' def}
\end{align}
Assuming discreteness in time, the goal of DMD is to find a linear operator $A$ such that
\begin{equation}
  X' \approx A X.
  \label{eq: dmd_operator}
\end{equation}
A common approach is to perform a reduced singular value decomposition (SVD) of $I$:
\begin{equation}
  X = U \Sigma V^H,
  \label{eq: svd}
\end{equation}
where $H$ denotes the Hermitian conjugate, $U\in\mathbb{C}^{n\times r}$, $\Sigma\in\mathbb{C}^{r\times r}$, and $V\in\mathbb{C}^{m\times r}$, with $r$ chosen, either as the full rank or a truncation rank. Projecting \autoref{eq: dmd_operator} onto the subspace spanned by $U$, we obtain the reduced operator
\begin{equation}
  \tilde{A} = U^H X' V \Sigma^{-1}.
  \label{eq:reducedA}
\end{equation}
The eigen-decomposition of $\tilde{A}$,
\begin{equation}
  \tilde{A} \tilde{W} = \tilde{W} \Lambda,
\end{equation}
yields eigenvalues $\Lambda$ and eigenvectors $\tilde{W}$. The DMD modes, which describe the spatial patterns, are then recovered as
\begin{equation}
  W = X' V \Sigma^{-1} \tilde{W}.
\end{equation}
The continuous version of operator A, called $\mathcal{A}$, is obtained via solving the linear differential equation that describes the dynamics;
\begin{equation}
    \frac{d \mathbf{I}}{dt} \approx \mathcal{A} \mathbf{I} \rightarrow \mathbf{I}(t + \Delta t) = e^{\mathcal{A} \Delta t} \mathbf{I}(t),
\end{equation}
where $\mathcal{A}$ is related to $A$ via
\begin{equation}
    A = e^{\mathcal{A} \Delta t},
\end{equation}
where $\Delta t$ quantifies the distance between consecutive frames. In our discussion of discrete dynamic mode decomposition above, we implicitly assumed $\Delta t = 1$. Each eigenvalue $\lambda_i$ in $\Lambda$ in the discrete-time DMD relates to the temporal evolution of the corresponding mode in the continuous-time DMD via
\begin{equation}
  \phi_i(t) = \exp(\Omega_i t), \quad \text{with } \Omega_i = \log(\Lambda_i),
  \label{eq: Lambda vs. Omega}
\end{equation}
where $\Omega = \alpha + i \omega$ captures growth/decay in the linear dynamical system through its real part $\alpha$, and captures oscillatory motion through its imaginary part $\omega$, which can be dubbed as angular frequency. This spectral decomposition will give us insight about the underlying dynamics governing the evolution of a dynamical system. 

\section{GRMHD Modal Analysis}
\begin{figure*}[t]
    \centering
    \includegraphics[width=1.0\linewidth]{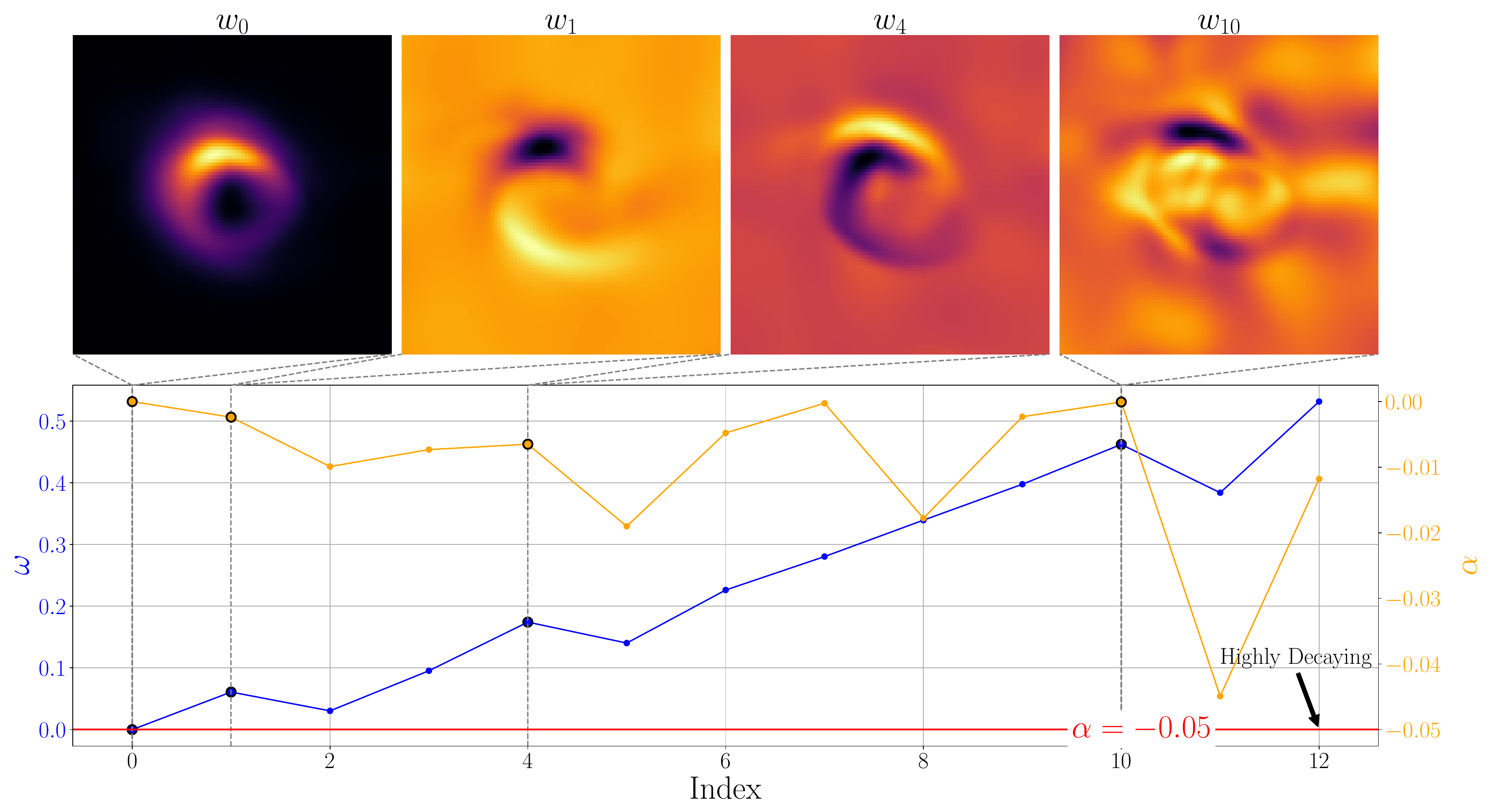}
    \caption{GRMHD modes and eigenvalues of \neuralDMD reconstruction from \ngehtp coverage. The top panel shows both the real (red) and imaginary (blue) parts of the eigenvalues. The imaginary part of the eigenvalues, $\omega = \text{Im}(\Omega)$, indicates the oscillatory behavior and the real part, $\alpha = \text{Re}(\Omega)$, indicates the decay rate. Note that the higher order modes have highly decaying behavior $\alpha < -0.05$, indicating that the reconstruction quality and interpretability of the dynamical system does not depend on these modes. None of the modes depicted here are highly decaying.}
    \label{fig: grmhd2 omegas_modes}
\end{figure*}

\begin{figure}[t]
    \centering
    \includegraphics[width=0.5\linewidth]{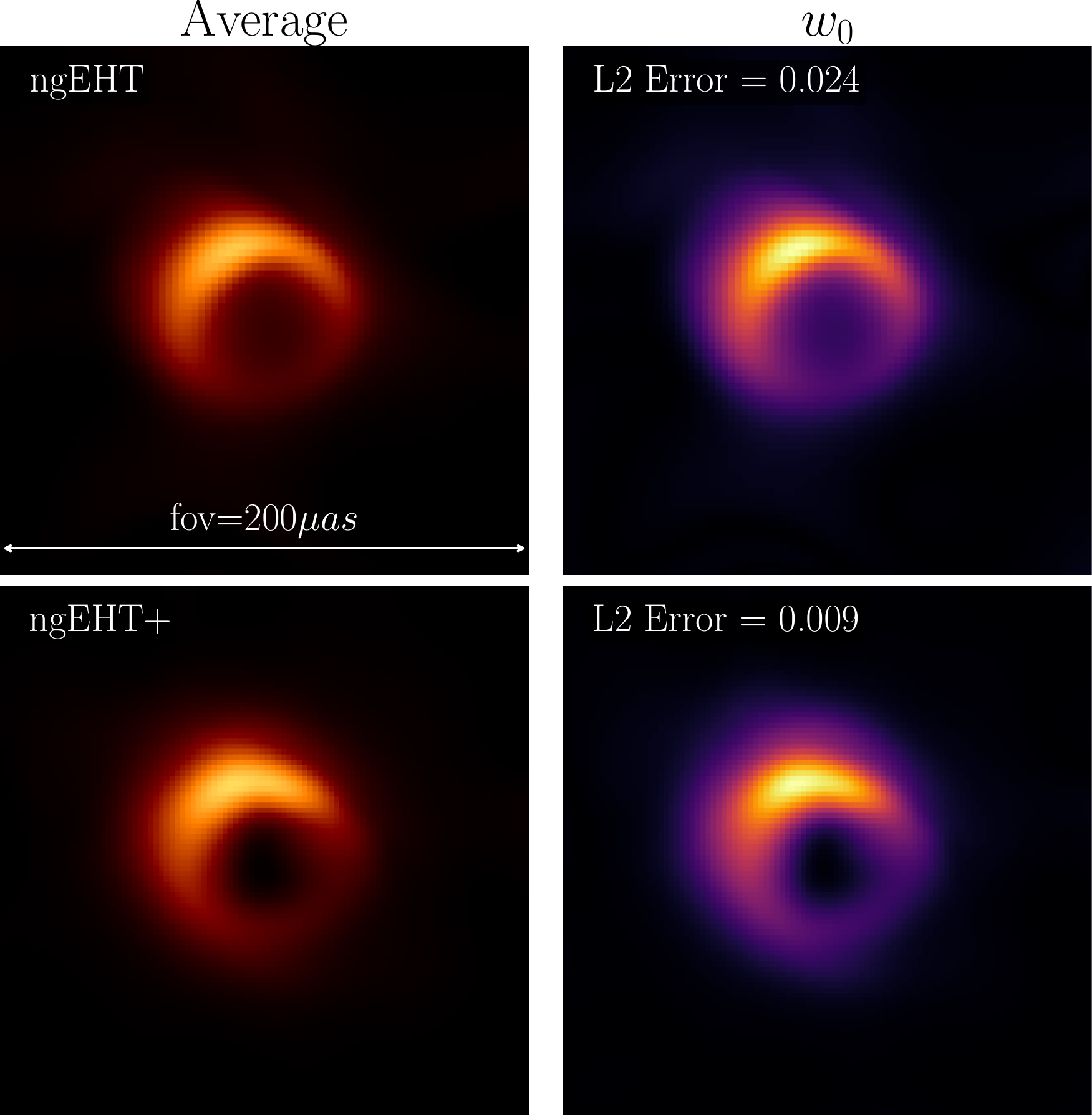}
    \caption{zeroth-order modes of the GRMHD experiment as compared to the average frames, for \ngeht and \ngehtp coverages. Note that there is but small differences between $w_0$ and the average.}
    \label{fig: grmhd2 mode0}
\end{figure}

One interesting aspect of DMD, and by extension \neuralDMD, is that the modes learned are interpretable. In GRMHD simulations, the learned modes as a result of fitting \neuralDMD to sparse observations are comprised of a static mode that captures the time-average of the dynamics within the data-fitting window, and a set of spiral features that capture the main dynamics. In \autoref{fig: grmhd2 omegas_modes} we show some of the GRMHD modes along with their spectrum. Moreover, in \autoref{fig: grmhd2 mode0} it is shown that empirically, the zeroth order mode learns the average frame.

\section{Additional Details: Image Domain}
\label{sec:supp_additional_experimental_details}

\subsection{Weather Data Assimilation}
\label{sec:supp_weather_da}

\paragraph{\textbf{ERA5 dataset and preprocessing.}}
We evaluate \neuralDMD on ERA5~\cite{era5_cds_2023,ERA5_paper}, a global weather reanalysis produced by ECMWF. We use hourly near-surface wind-speed magnitudes at $10$m above the surface over North America for April $1$--$7$, $2025$ (168 frames). Ground-truth frames are available on the ERA5 grid and are used for evaluation on the full spatial domain (dense grid). The wind-speed magnitude is computed from its longitude $u$ and latitude $v$ components, so that $\text{magnitude} = \sqrt{u^2 + v^2}$.

\paragraph{\textbf{Sparse observation model.}}
To mimic sparse station measurements, we restrict observations in each frame to a set of sensor coordinates. Observation locations are derived from the World Meteorological Organization (WMO) OSCAR/Surface archive as part of WIGOS~\cite{WMO_WIGOS_Guide_2019,WMO_OSCAR_Surface}. We retain stations labeled ``Operational'' and use their reported geographic coordinates as sensor sites. In the main experiment, we use $\sim 5000$ station observations per frame (Fig.~\ref{fig: weather baseline}).
We map station locations to the ERA5 grid via bilinear interpolation to obtain the corresponding sparse measurements $I(x_k,y_k,t)$.

\paragraph{\textbf{Baseline: minimalist 3D-Var / \varbaseline.}}
We compare against a minimalist implementation of three-dimensional variational data assimilation (\varbaseline), designed to avoid strong forecast-model priors. Classical 3D-Var combines a background forecast with observations by minimizing a weighted least-squares objective with background/observation error covariances~\cite{Lorenc1986,Courtier1998}. In our implementation, we use a static background equal to the initial frame and assume uncorrelated Gaussian background and observation errors with uniform variance. We compute kernel-smoothed innovations (observation minus background) at station locations and apply a single scalar weight in an optimal-interpolation-style analysis~\cite{OI_method,OI_vs_3DVAR}. Observations are mapped to the grid using a nearest-gridpoint observation operator (each station is associated with the closest grid cell in latitude and longitude), and we assume diagonal background and observation covariances with uniform variances \( \sigma_b = 2.0\,\mathrm{m\,s^{-1}} \) and \( \sigma_o = 1.0\,\mathrm{m\,s^{-1}} \). The resulting quadratic 3D-Var objective is minimized at each timestep using L-BFGS-B with a maximum of 50 iterations.

\paragraph{\textbf{INR baselines (vanilla MLP, SIREN).}}
For image-domain baselines, we use two spatio-temporal coordinate networks that directly regress $I(x,y,t)$ from $(x,y,t)$: a vanilla MLP coordinate network and SIREN~\cite{siren}. For fairness, INR baselines use the same backbone as \neuralDMD's mode network, with time included as an additional input and a comparable positional-encoding bandwidth. We select the positional-encoding bandwidth from the same range used elsewhere in the paper (frequencies up to $L\in[0,10]$) and use smaller bandwidth under higher sparsity. We choose the best $L$ with trial and error.

\paragraph{\textbf{\neuralDMD configuration for weather.}}
We train \neuralDMD using the image-domain objective in Eq.~\eqref{eq:neuraldmd_direct_loss}, where spatial coordinates correspond to longitude and latitude. The number of modes is chosen based on sparsity (typically $r\in[5,75]$); additional ablations on mode count versus sparsity are provided in Sec.~\ref{subsec:supp_weather_ablations_modes}. Unless otherwise stated, parameters are randomly initialized.

\paragraph{\textbf{Nowcasting / forecast protocol.}}
To evaluate short-term forecasting, we hold out April $8$, $2025$ (the day following the training period) as ground truth for predictive evaluation. We report reconstruction metrics within the data-fitting window and forecasting metrics on held-out future frames. The forecasting horizon considered is 5 hours.

\paragraph{\textbf{Evaluation metrics.}}
We report RMSE, PSNR, SSIM, and LPIPS computed against dense ERA5 ground-truth frames. Metrics are computed per frame over the full grid and then averaged over time; we report mean $\pm$ standard deviation across frames. LPIPS uses AlexNet backbone.

\paragraph{\textbf{Additional qualitative results and robustness.}}
Additional comparisons under varying sparsity levels and additional qualitative frames (including forecast examples beyond the observation window) are provided in Sec.~\ref{sec: robustness experiments}. 

\begin{figure}
    \centering
    \includegraphics[width=0.5\linewidth]{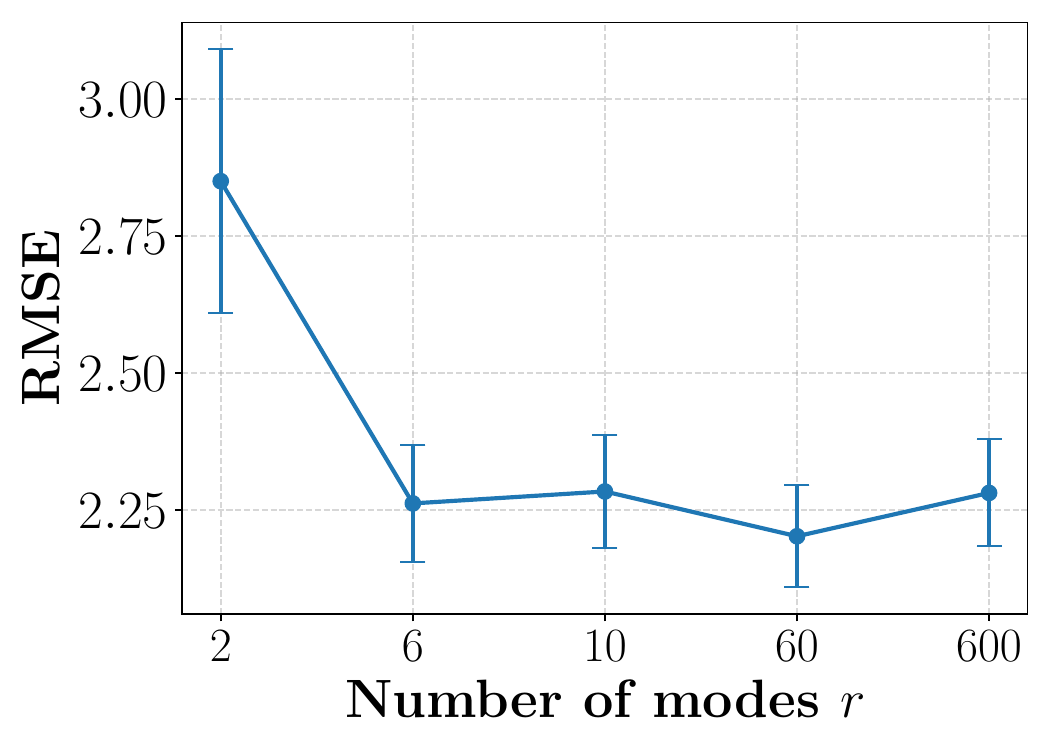}
    \caption{\neuralDMD reconstruction error with varying number of modes. Note that the only high-RMSE case is $r=2$ which means that using only 2 modes is not enough for high-fidelity reconstruction.}
    \label{fig:weather mode ablation}
\end{figure}

\paragraph{\textbf{Weather Ablations: Error vs. Mode Count}}
\label{subsec:supp_weather_ablations_modes}
We provide an ablation study analyzing how performance varies with the number of modes $r$ under the observation sparsity level. In practice, performance is relatively insensitive beyond a moderate $r$, as unused modes are driven to fast decay. Fig.~\ref{fig:weather mode ablation} shows that varying number of modes does not change the RMSE of the \neuralDMD prediction beyond a small amount.

\subsection{Wave Equation: Simulation-Trained vs.\ Untrained Modeling}
\label{sec:supp_wave_coral}

\paragraph{\textbf{Goal.}}
This experiment isolates the effect of distribution shift on simulation-trained operator-learning models in sparse inverse settings. We compare CORAL~\cite{CORAL_INR} (trained on simulations) to \neuralDMD (fit per-sequence directly to sparse observations).

\paragraph{\textbf{Data generation.}}
We consider the 2D wave equation
\begin{equation}
\frac{\partial^2 u(x,y,t)}{\partial t^2} = c \nabla^2 u(x,y,t),
\end{equation}
and generate sequences using an analytic modal form
\begin{equation}
u(x,y,t) = \sum_{k=1}^{R} \sin(m_k \pi x)\sin(n_k \pi y)\cos(\omega_{m_kn_k} t), \quad
\omega_{mn}=c\sqrt{(m\pi)^2+(n\pi)^2}.
\end{equation}
We fix $c=1$ and $R=3$. Training sequences are generated by sampling $m_k,n_k \sim \mathrm{Unif}(\{1,\dots,4\})$ for each mode $k$, producing $N_{\mathrm{train}}=5000$ sequences. For evaluation, we generate (i) an in-distribution (ID) test set with $m_k,n_k \in [1,4]$ and (ii) an out-of-distribution (OOD) test set with $m_k,n_k \in [10,16]$ (higher spatial frequencies).

\paragraph{\textbf{Sparse observation model.}}
For each sequence, we assume access only to sparse samples of the spatio-temporal field. Let $\mathcal{S}=\{(x_i,y_i,t_i)\}_{i=1}^{N_{\mathrm{obs}}}$ denote the observed coordinate set.
We use uniform random sampling over space-time with $N_{\mathrm{obs}}=$ 1474 $\times$ 25 observations total. Sampling is done per-frame, the spatial locations are varied per-frame, and no additional noise is added. 

\paragraph{\textbf{Baselines.}}
\textbf{CORAL~\cite{CORAL_INR}.} We train CORAL on the $N_{\mathrm{train}}=5000$ simulated training sequences described above. CORAL is provided sparse frame observations and outputs a continuous field prediction $u(x,y,t)$. During inference, all the weights of CORAL are frozen and only the latent code is optimized so that the INR autodecoder is able to recover the frames. The pretrained Neural ODE is then used to predict dynamics.

\textbf{\neuralDMD.} We fit \neuralDMD per-sequence by minimizing the measurement residual over the same observation set $\mathcal{S}$ used for CORAL, using the image-domain objective (Eq.~\ref{eq:neuraldmd_direct_loss}). Hyperparameters follow Sec.~\ref{subsec:setup} with [$r = 40$ modes, positional encoding $L = 8$] for this experiment.

\paragraph{\textbf{Evaluation protocol and metrics.}}
We evaluate reconstructions on the full dense grid against ground-truth frames (not only at observed points). We report [RMSE/PSNR/SSIM] averaged over [16 test sequences] for both ID and OOD settings at sparsity levels of $10\%$ and $40\%$, respectively.
\begin{figure}
    \centering
    \includegraphics[width=0.65\linewidth]{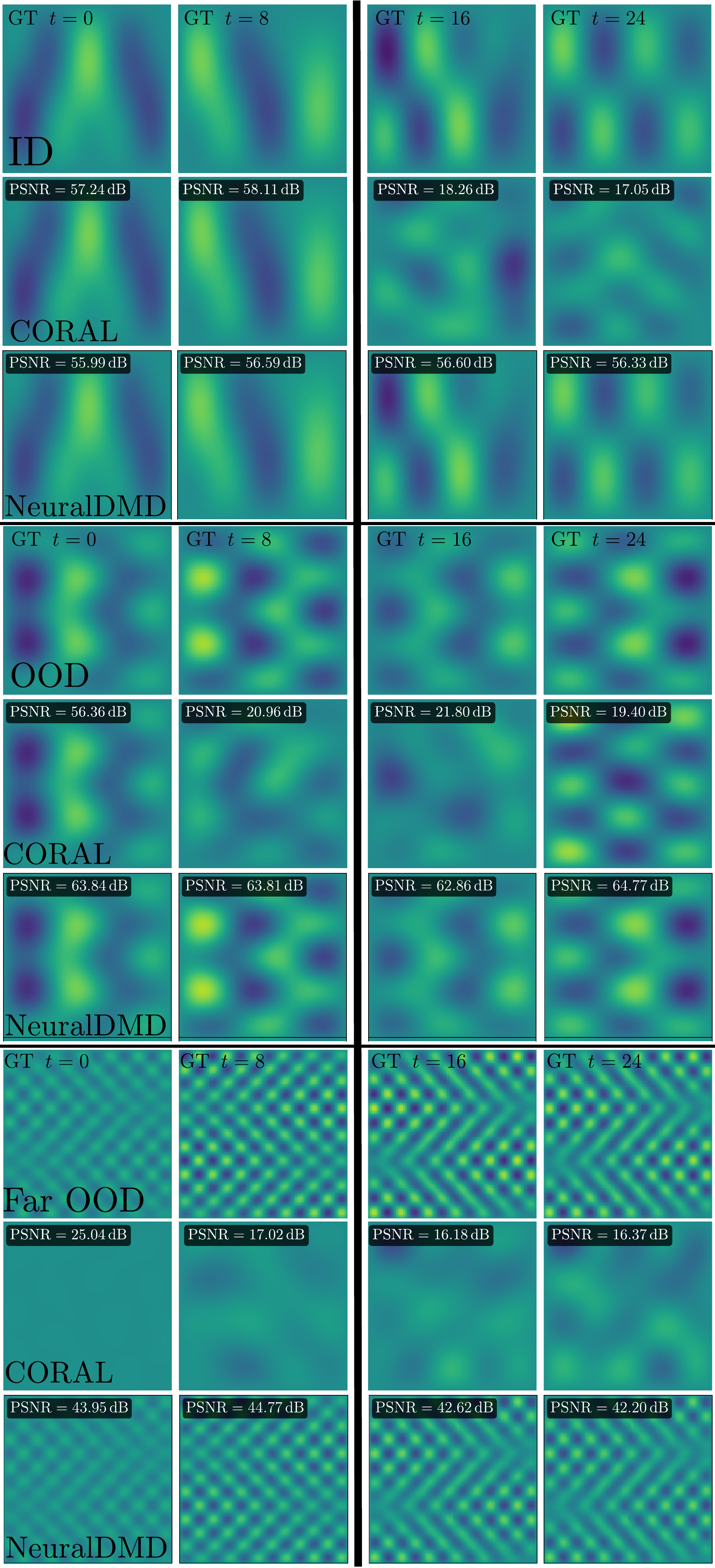}
    \caption{CORAL and \neuralDMD reconstructions of ID, OOD, and far OOD 2d wave equation test cases. \neuralDMD outperforms CORAL on all cases, most notably when the test case is far outside of the distribution CORAL is trained on. The left two frames for CORAL and \neuralDMD at times $t = 0, 8$ are reconstructed from 10$\%$ uniformly sampled pixels for ID case and $40\%$ samples for OOD and far OOD case. The right two frames $t = 16, 24$ are then extrapolated with no observational data.}
    \label{fig:wave_equation}
\end{figure}

\paragraph{\textbf{Results.}}
Qualitatively, CORAL is expected to perform well in-distribution but can degrade under frequency shift, often producing overly smooth (low-pass) reconstructions when trained only on low-frequency waves, while \neuralDMD remains applicable because it is optimized directly to each sequence’s measurements. Fig.~\ref{fig:wave_equation} provides qualitative results for ID (wave lengths in range of training data), OOD (wave lengths smaller than training data), and far OOD (wave lenghts much smaller than training data) cases for CORAL and \neuralDMD. \neuralDMD is capable of reconstructing ID, OOD, and far OOD while CORAL struggles with OOD and fails on far OOD.

\section{Additional Details: Fourier-Domain (EHT/GRMHD) Experiments}
\label{sec:supp_eht}

\subsection{GRMHD Simulations and Image-Domain Ground Truth}
\label{sec:supp_grmhd}

\paragraph{\textbf{GRMHD simulations.}}
We evaluate on physically realistic black hole accretion simulations generated with General Relativistic Magnetohydrodynamics (GRMHD)~\cite{ipol_paper,H-ARM_Gammie_2003,grmhd_survey}. These simulations model magnetized accretion flows in strong gravitational fields and produce time-varying plasma dynamics.

\paragraph{\textbf{Ray tracing to the image domain.}}
We render image-plane sequences from the GRMHD flow using a general-relativistic ray-tracing pipeline~\cite{H-ARM_Gammie_2003,grmhd_survey}, producing ground-truth frames $I_t(x,y)$ as seen by a distant observer. In the main paper we show a near face-on configuration; additional results for an edge-on configuration are provided in Sec.~\ref{sec: edge-on grmhd}.

\paragraph{\textbf{Interstellar scattering.}}
To mimic Galactic-center observing conditions, we apply a scattering screen that models interstellar broadening~\cite{Johnson_2018,Issaoun_2021}. We use the scattering model implemented in the \texttt{stochastic\_optics} module of \ehtim.

\subsection{Simulating EHT 2017 Interferometric Observations}
\label{sec:supp_eht_obs}

\paragraph{\textbf{Sparse Fourier sampling and uv-coverage.}}
EHT observations provide sparse, time-varying samples of the Fourier transform of the sky brightness. We simulate EHT-like measurements using \ehtim~\cite{eht-imaging}, which models Earth-rotation synthesis and measurement noise/corruptions~\cite{Chael_2016,Chael_2018,eht-imaging}. We generate uv-coverage using \ngehtsim~\cite{Pesce_ngehtsim_2023} to mimic the observing geometry of the April 11, 2017 campaign on Sgr~A$^\ast$ (\ehtobs;~\cite{sgrA_paper1}). The resulting measurement set consists of complex visibilities $\{V_i\}$ observed at coordinates $\{(u_i,v_i,t_i)\}$ with associated uncertainty estimates $\{\sigma_i\}$. We provide detailed time-dependent sampling patterns in Sec.~\ref{sec: coord tables and patterns}.

\begin{figure}
    \centering
    \includegraphics[width=\linewidth]{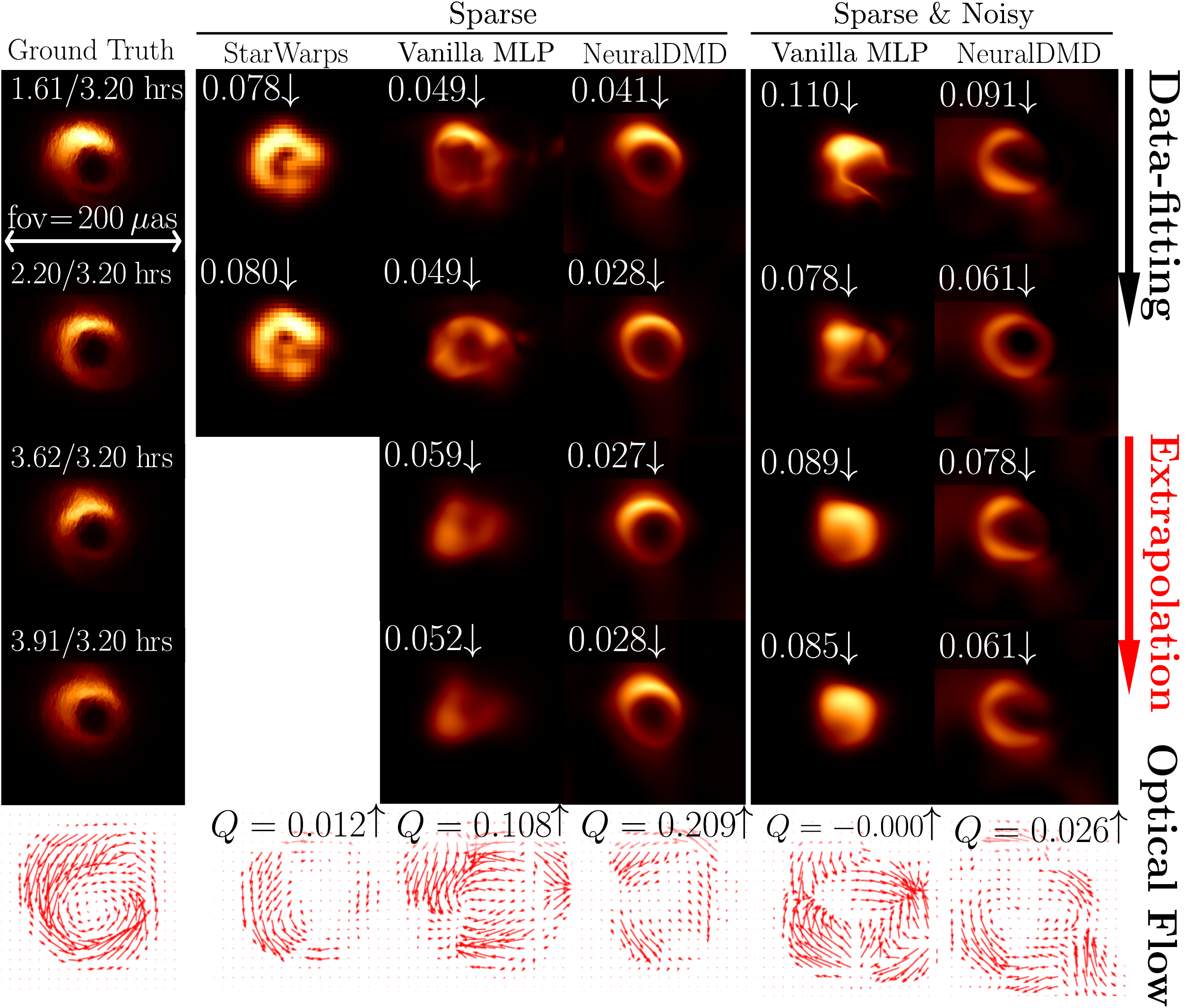}
    \caption{GRMHD reconstruction from sparse interferometric data with and without additional gain corruptions.}
    \label{fig:gain_results}
\end{figure}

\paragraph{\textbf{Noise and station-based corruptions.}}
We consider (i) thermal (Gaussian) noise and (ii) station-based atmospheric corruptions (complex gains) affecting each telescope. Unless otherwise stated, the simulated measurement uncertainties $\sigma_i$ are those returned by the \ehtim/\ngehtsim simulation pipeline. Additional simulation details and parameter settings are provided in Sec.~\ref{sec: coord tables and patterns}. Fig.~\ref{fig:gain_results} shows the results for \neuralDMD and some baselines for cases of sparse data, and sparse and noisy data (with the additional gain corruptions).

\subsection{Fitting Objectives and Gain Marginalization}
\label{sec:supp_eht_fitting}

\paragraph{\textbf{Visibility-domain data term.}}
For Fourier-domain measurements, we fit models by matching predicted and observed visibilities. For experiments where per-measurement uncertainties are available, we use the reduced chi-squared objective
\begin{equation}
\chi^2(\Theta)=\frac{1}{N}\sum_{i=1}^{N}\frac{|V_i-\hat V_i(\Theta)|^2}{\sigma_i^2},
\label{eq:supp_chisq}
\end{equation}
where $\hat V_i(\Theta)$ denotes the model-predicted visibility at $(u_i,v_i,t_i)$ and $N$ is the number of observations.

\paragraph{\textbf{Gain marginalization.}}
Station gains distort both visibility amplitudes and phases, so directly optimizing Eq.~\eqref{eq:supp_chisq} can bias reconstructions. We therefore marginalize over gains following prior work (see \cite{THEMIS} and Sec.~\ref{sec: gain marginalization} for derivations and implementation). We choose marginalization instead of closure-only objectives to avoid degeneracies and unconstrained modes that can arise when using only gain-invariant quantities~\cite{EHTIII2019,Fish2016,Chael2018}.

\subsection{Baselines and Initialization}
\label{sec:supp_eht_baselines}

\paragraph{\textbf{Spatio-temporal INR baseline (vanilla MLP).}}
In all Fourier-domain experiments, the ``neural representation'' baseline corresponds to the spatio-temporal INR baseline implemented as a vanilla MLP coordinate network that regresses $I(x,y,t)$ from $(x,y,t)$ and is optimized directly to the visibilities using the same Fourier measurement loss.

\paragraph{\textbf{Matched initialization for fairness.}}
For fairness under extreme sparsity, \neuralDMD and the vanilla MLP INR baseline use the same structured initialization and random seed. Specifically, we pretrain the vanilla MLP INR on a constant disk of the expected source size and initialize \neuralDMD spatial modes using orthogonal Zernicke polynomials to provide a diverse multi-scale starting basis. Additional initialization variants (including random initialization and alternatives for denser coverage) are reported in Sec.~\ref{sec:ngeht_more_init}.

\begin{figure}
    \centering
    \includegraphics[width=1.0\linewidth]{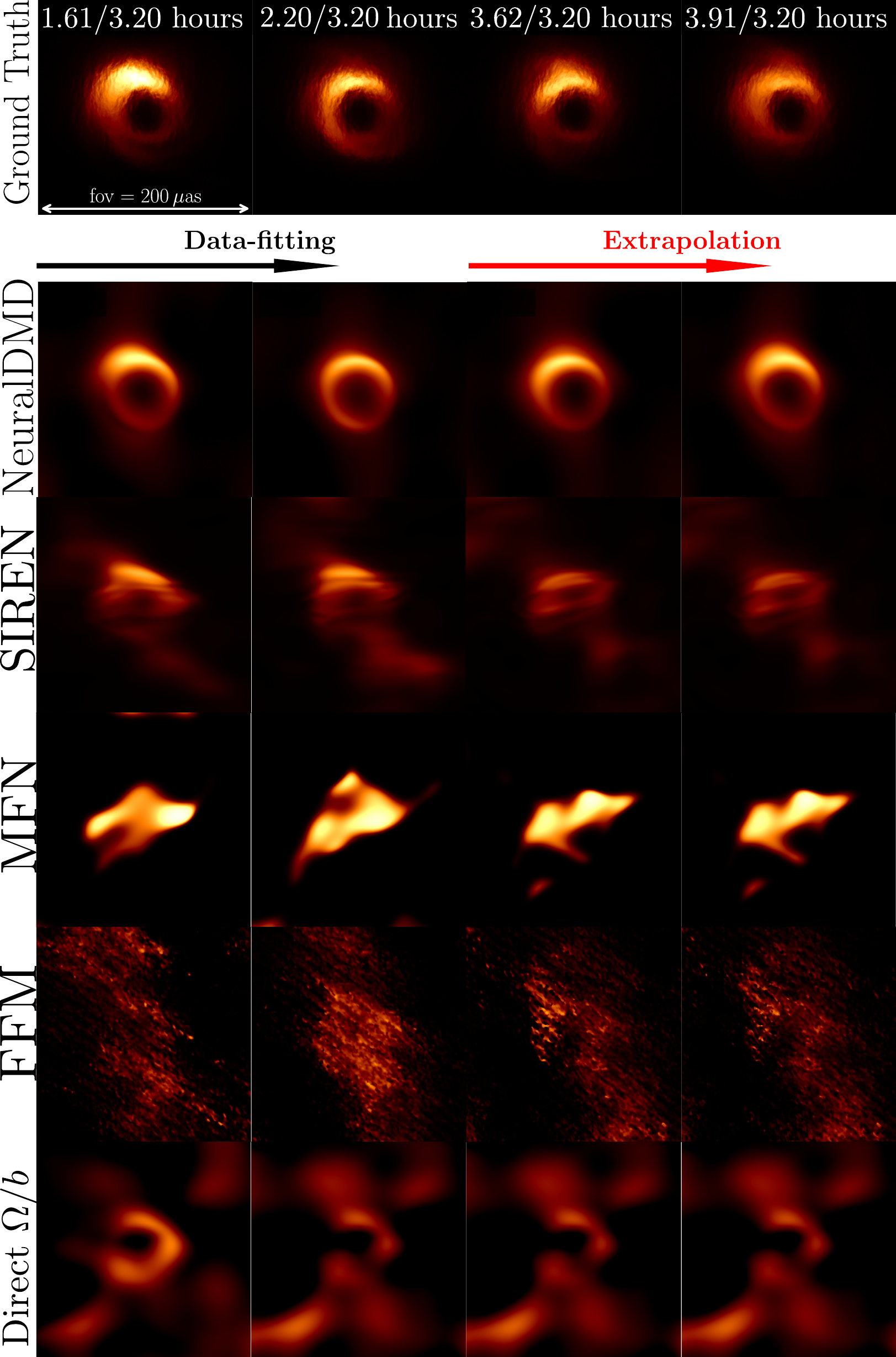}
    \caption{\neuralDMD reconstruction results as compared to additional INR baselines along with \neuralDMD but with direct optimization on $\Omega, b$ parameters. \neuralDMD outperforms all baselines.}
    \label{fig:inr_qualitative_supp}
\end{figure}

\begin{table}[t]
\centering
\renewcommand{\arraystretch}{1.2}
\caption{Comparison of $\chi^2$ values across different model architectures for black hole EHT 2017 experiment.}
\setlength{\tabcolsep}{8pt}
\begin{tabular}{lcccccc}
\toprule
Metric & Vanilla MLP & SIREN & MFN & FFM & Direct $\Omega/b$ & NeuralDMD \\
\midrule
$\chi^2$ & 0.92 & 0.94 & 0.86 & 0.7 & 1.24 & 1.09 \\
\bottomrule
\end{tabular}
\label{tab:chi2_models}
\end{table}

\begin{table}[t]
\centering
\caption{Ablation study across architectures and frequency choices.}
\label{tab:quantitative}
\begin{tabular}{lccccc}
\toprule
Method & Freq & RMSE $\downarrow$ & PSNR $\uparrow$ & SSIM $\uparrow$ & LPIPS $\downarrow$ \\
\midrule

\siren & 1  & 0.1$\pm$0.02 & 20$\pm$1 & 0.5$\pm$0.02 & 0.4$\pm$0.03 \\
      & 3  & 0.07$\pm$0.01 & 20$\pm$1 & 0.6$\pm$0.03 & 0.3$\pm$0.03 \\
      & 10 & 0.08$\pm$0.006 & 20$\pm$0.7 & 0.6$\pm$0.02 & 0.3$\pm$0.01 \\
\midrule

\mfn & 1  & 0.1$\pm$0.007 & 20$\pm$0.5 & 0.5$\pm$0.01 & 0.3$\pm$0.006 \\
     & 3  & 0.1$\pm$0.01 & 20$\pm$0.7 & 0.5$\pm$0.009 & 0.3$\pm$0.004 \\
     & 10 & 0.2$\pm$0.03 & 20$\pm$2 & 0.5$\pm$0.02 & 0.3$\pm$0.02 \\
\midrule

\ffm & 64   & 0.2$\pm$0.06 & 20$\pm$4 & 0.3$\pm$0.1 & 0.8$\pm$0.2 \\
     & 256  & 0.09$\pm$0.004 & 20$\pm$0.3 & 0.4$\pm$0.006 & 0.7$\pm$0.02 \\
     & 1024 & 0.09$\pm$0.005 & 20$\pm$0.4 & 0.5$\pm$0.03 & 0.6$\pm$0.02 \\
\midrule

Vanilla MLP & 1  & 0.09$\pm$0.02 & 20$\pm$2 & 0.7$\pm$0.03 & 0.3$\pm$0.02 \\
            & 3  & 0.05$\pm$0.003 & 30$\pm$0.6 & 0.8$\pm$0.01 & 0.2$\pm$0.004 \\
            & 10 & 0.2$\pm$0.1 & 20$\pm$6 & 0.6$\pm$0.08 & 0.3$\pm$0.04 \\
\midrule

Direct $\Omega/b$
& 1  & 0.09$\pm$0.01 & 20$\pm$1 & 0.4$\pm$0.04 & 0.2$\pm$0.03 \\
& 2  & 0.1$\pm$0.004 & 20$\pm$0.3 & 0.3$\pm$0.01 & 0.4$\pm$0.03 \\
& 10 & 0.2$\pm$0.004 & 10$\pm$0.2 & 0.2$\pm$0.0002 & 0.6$\pm$0.01 \\
\midrule

\textbf{\neuralDMD}
& \textbf{1}  & \textbf{0.07$\pm$0.007} & \textbf{20$\pm$1} & \textbf{0.6$\pm$0.02} & \textbf{0.2$\pm$0.02} \\
& \textbf{2} & \textbf{0.03$\pm$0.01} & \textbf{30$\pm$3} & \textbf{0.7$\pm$0.3} & \textbf{0.2$\pm$0.08} \\
& \textbf{10} & \textbf{0.07$\pm$0.003} & \textbf{20$\pm$0.5} & \textbf{0.5$\pm$0.002} & \textbf{0.3$\pm$0.04} \\
\bottomrule
\end{tabular}
\end{table}
\paragraph{\textbf{Additional Quantitative Results.}} In the main text, Tab.~\ref{tab:eht_ablation_compact} shows quantitative results of \neuralDMD and the baselines for two choices of number of input frequencies. Tab.~\ref{tab:quantitative} shows a more complete result for three input frequencies choices.

\paragraph{\textbf{Other Baselines Qualitative Results.}} In Fig.\ref{fig: combined_eht2017_all} we provided qualitative results of \neuralDMD, and vanilla MLP and \sw baselines only. Fig.~\ref{fig:inr_qualitative_supp} shows \neuralDMD against additional INR baselines and \neuralDMD when instead of optimizing neural network parameters $\Theta_\Omega, \Theta_b$, the parameters $\Omega, b$ are directly optimized. These are the remaining baselines whose quantitative results are show in Tab.~\ref{tab:eht_ablation_compact}; Fig.~\ref{fig:inr_qualitative_supp} shows the best-case result for each baseline (which is the low frequency choice in Tab.~\ref{tab:eht_ablation_compact}). Furthermore, the $\chi^2$ value reached at the end of training for the best-case result are shown in Tab.~\ref{tab:chi2_models}. Note that the table shows some representative values and are not a good quantity for comparing the baselines. All $\chi^2$ reached for all fits in this paper are $~1.0$ (there are no bad fits to the data).

\subsection{\optdmd Baseline with TV Regularization}
\label{sec:supp_optdmd_tv}

\paragraph{\textbf{Why TV regularization.}}
When constructing an \optdmd baseline for comparison under sparse, noisy visibilities, we include a total-variation (TV) regularizer on reconstructed image frames to promote spatial smoothness and stabilize convergence. Full details of the \optdmd+TV objective, hyperparameters, and optimization procedure are provided in this section.

\subsection{Robustness to Noise and Sparsity}
\label{sec:supp_eht_robustness}

We evaluate robustness across a grid of (i) visibility noise levels and (ii) Fourier sampling densities (number of visibilities per frame). We report qualitative reconstructions and quantitative errors for \neuralDMD and key baselines, including the vanilla MLP INR and \optdmd+TV. The complete noise--sparsity sweep and additional examples are provided in Sec.~\ref{sec: robustness experiments}.

\subsection{Additional Ablations}
\label{sec:supp_eht_ablations}

\paragraph{\textbf{Encoding bandwidth and INR backbone sweeps.}}
We sweep positional-encoding bandwidths $L\in[0,10]$ (and additional high-bandwidth settings for Fourier-feature baselines) and report the impact on reconstruction metrics. We also ablate INR backbones (SIREN, MFN, Fourier-feature MLP, and vanilla MLP) under matched optimization settings (Table~\ref{tab:eht_ablation_compact} and extended results in the supplement).

\paragraph{\textbf{Direct optimization of $(\Omega,b)$.}}
To isolate the effect of parameterizing the spectrum and initial coefficients with networks $\Theta_\Omega$ and $\Theta_b$, we compare to a variant that directly optimizes $\{\Omega_j,b_j\}$ as free variables while keeping the mode network and measurement loss fixed. This variant is substantially less stable under sparse visibilities and yields worse reconstructions (Table~\ref{tab:eht_ablation_compact}).

\section{Initialization}
\label{sec:ngeht_more_init}
\begin{figure*}[t]
    \centering
    \includegraphics[width=1.0\linewidth]{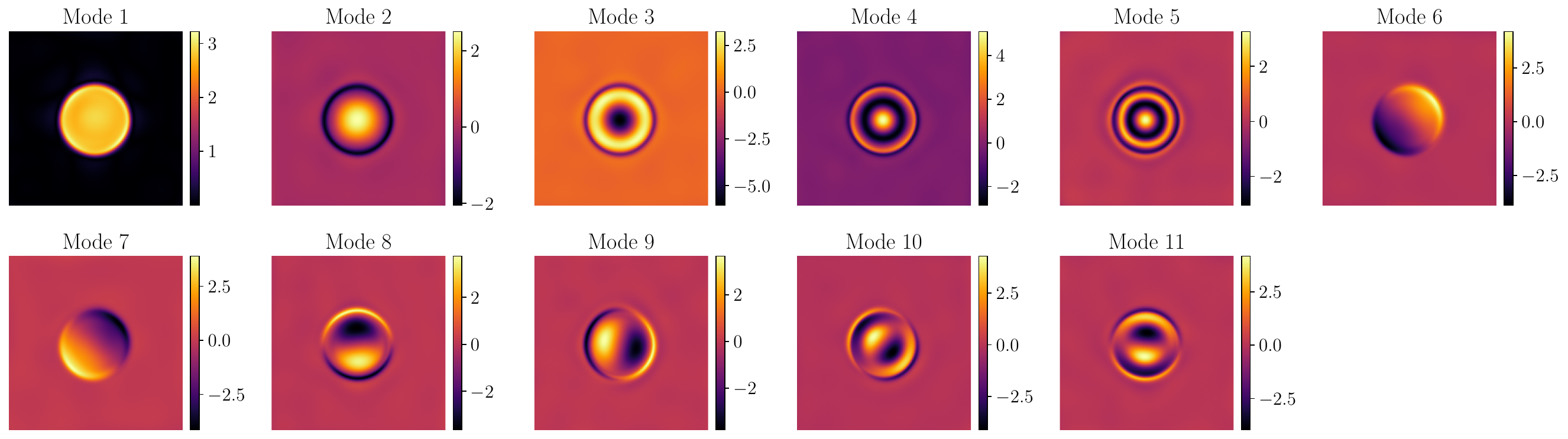}
    \caption{Top 11 Zernicke polynomials used for pretraining \neuralDMD. The ``disk'' initialization for neural representation is equivalent to the first Zernicke polynomial used for mode 1 of \neuralDMD.}
    \label{fig: Zernicke}
\end{figure*}
In the main text, two initializations for better conditioning were employed: 1. a disk of proportional size to the black hole for the neural representation baseline, and 2. Zernicke polynomials for \neuralDMD. Since we only use 10 dynamic modes + static, the modes are pretrained on the top 11 Zernicke polynomials, leading to the modes depicted in \autoref{fig: Zernicke}.

\section{Linear Dynamics for Black Holes}
\begin{figure*}
    \centering
    \includegraphics[width=1.0\linewidth]{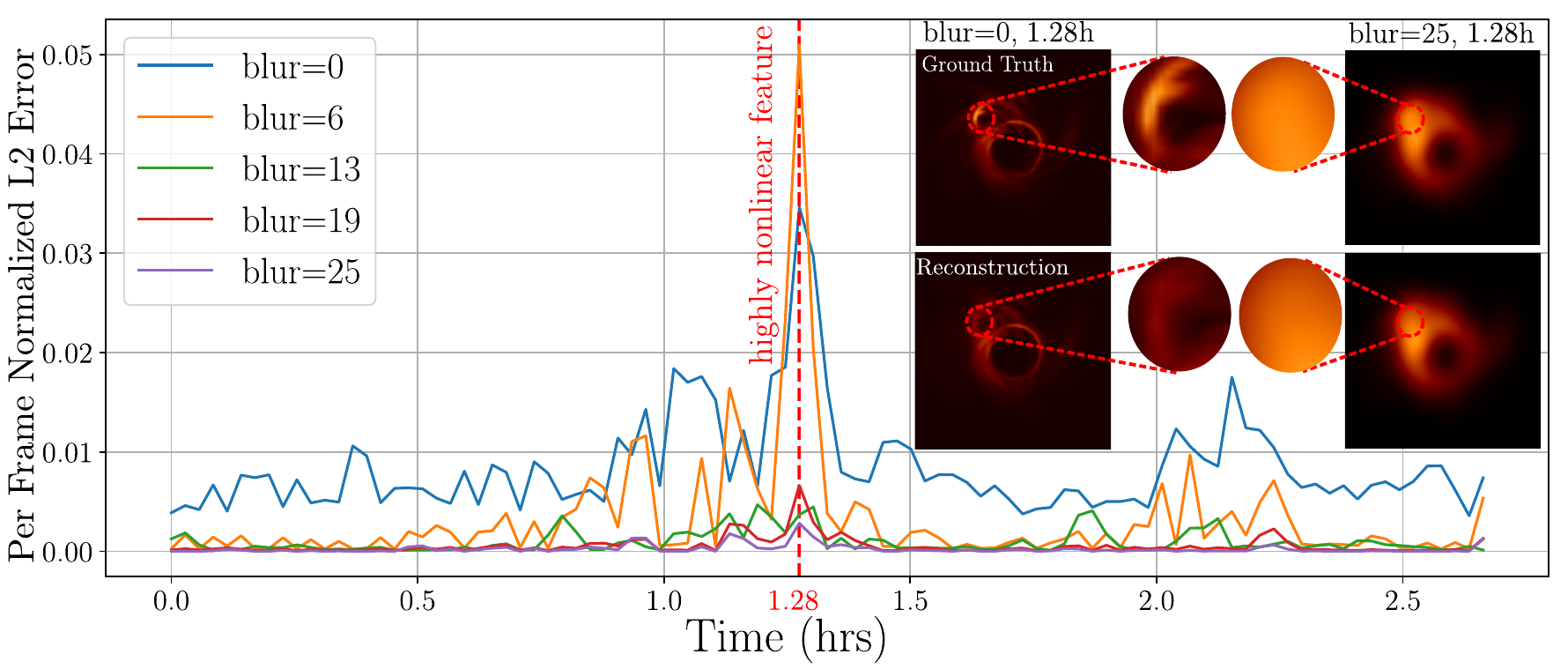}
    \caption{This figure demonstrates that the dynamics observed by the EHT, when blurred with a $25\mu$as beam, are well approximated by a linear dynamical system. The inset displays the frame at $1.28$ hours after the initial time: without blurring (blur $= 0$ column), highly nonlinear structures are present and not captured by \neuralDMD, while under blurring, these features are smoothed out. The nonlinear structure is outlined and shown in the zoomed-in panel. The curves show the per-frame normalized $\mathrm{L}_2$ error between the \neuralDMD reconstructions and the ground truth, with error decreasing as blurring increases and the nonlinear features are progressively suppressed.}
    \label{fig: blurring exp}
\end{figure*}

Although Dynamic Mode Decomposition (DMD) only fits a linear model, it is a principled surrogate for the nonlinear Koopman operator and can approximate the true dynamics arbitrarily well. For Event Horizon Telescope (EHT) data the linear assumption is even more compelling: at the EHT’s angular resolution, the intrinsic image is blurred enough that small-scale nonlinear structures are washed out. As Figure \ref{fig: blurring exp} shows, increasing the blur to the EHT beam steadily lowers the reconstruction error, indicating that the observable dynamics are effectively linear and can be captured by our \neuralDMD model.

\section{Expansion of the EHT 2017 Coverage}
\label{sec: ngeht and ngehtp}
\begin{figure*}
    \centering
    \includegraphics[width=0.9\linewidth]{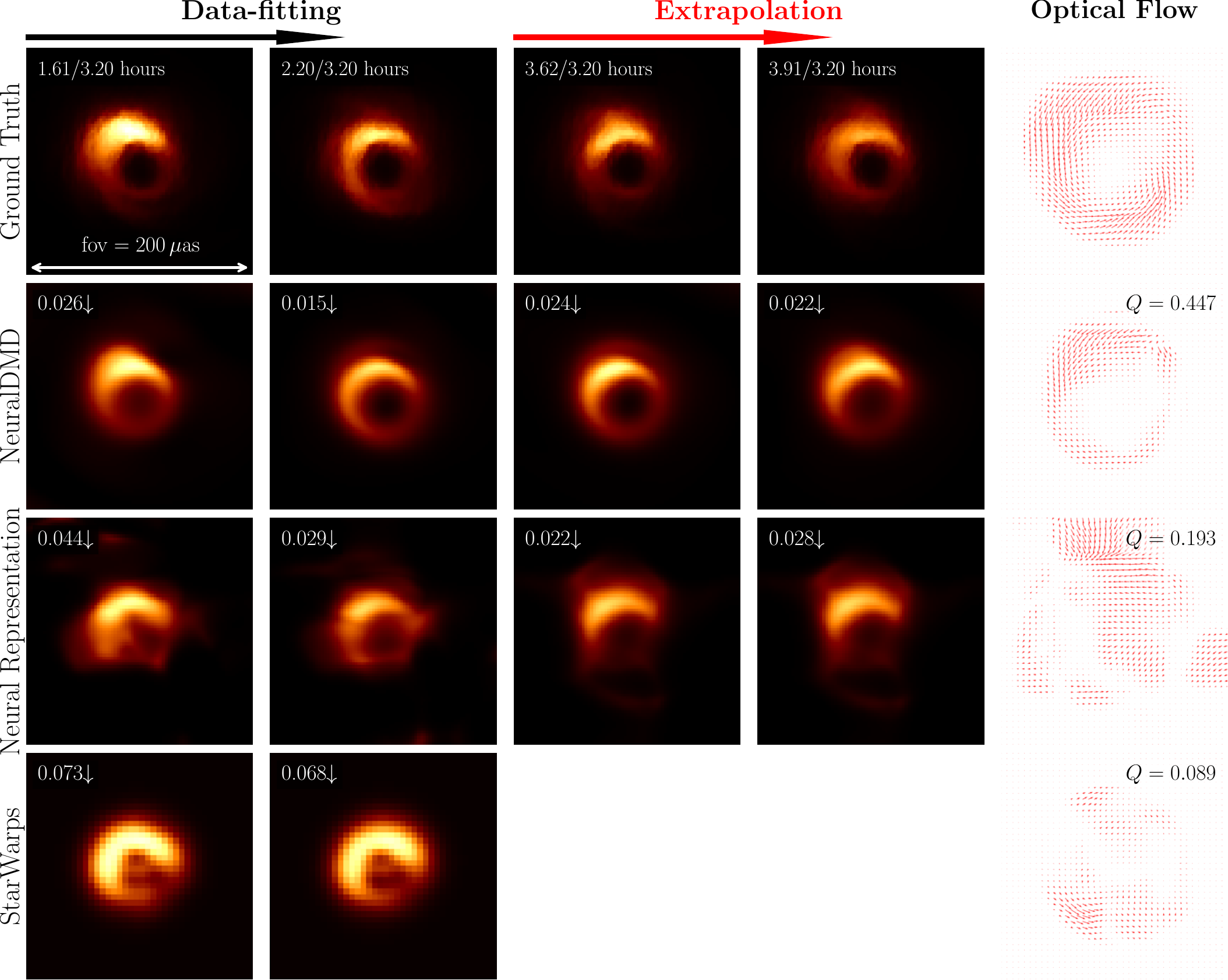}
    \caption{Reconstruction of a GRMHD sequence from \ngeht‑synthetic interferometric data. Rows (top to bottom): ground truth, \neuralDMD, neural representation, and \sw. The left two frames are data‑constrained reconstructions (``data-fitting''), the right two are model extrapolations. Numbers at each frame’s top‑left give the normalized L2 error versus ground truth (lower is better). \neuralDMD shows the lowest errors and best visual match across both observed and predicted frames. Since \sw directly fits pixels to the observation data, it cannot extrapolate The final column presents mean optical flow; $Q$ is the cosine similarity with the ground‑truth flow (higher is better).}
    \label{fig: GRMHD2 tests}
\end{figure*}

\begin{figure*}
    \centering
    \includegraphics[width=1.0\linewidth]{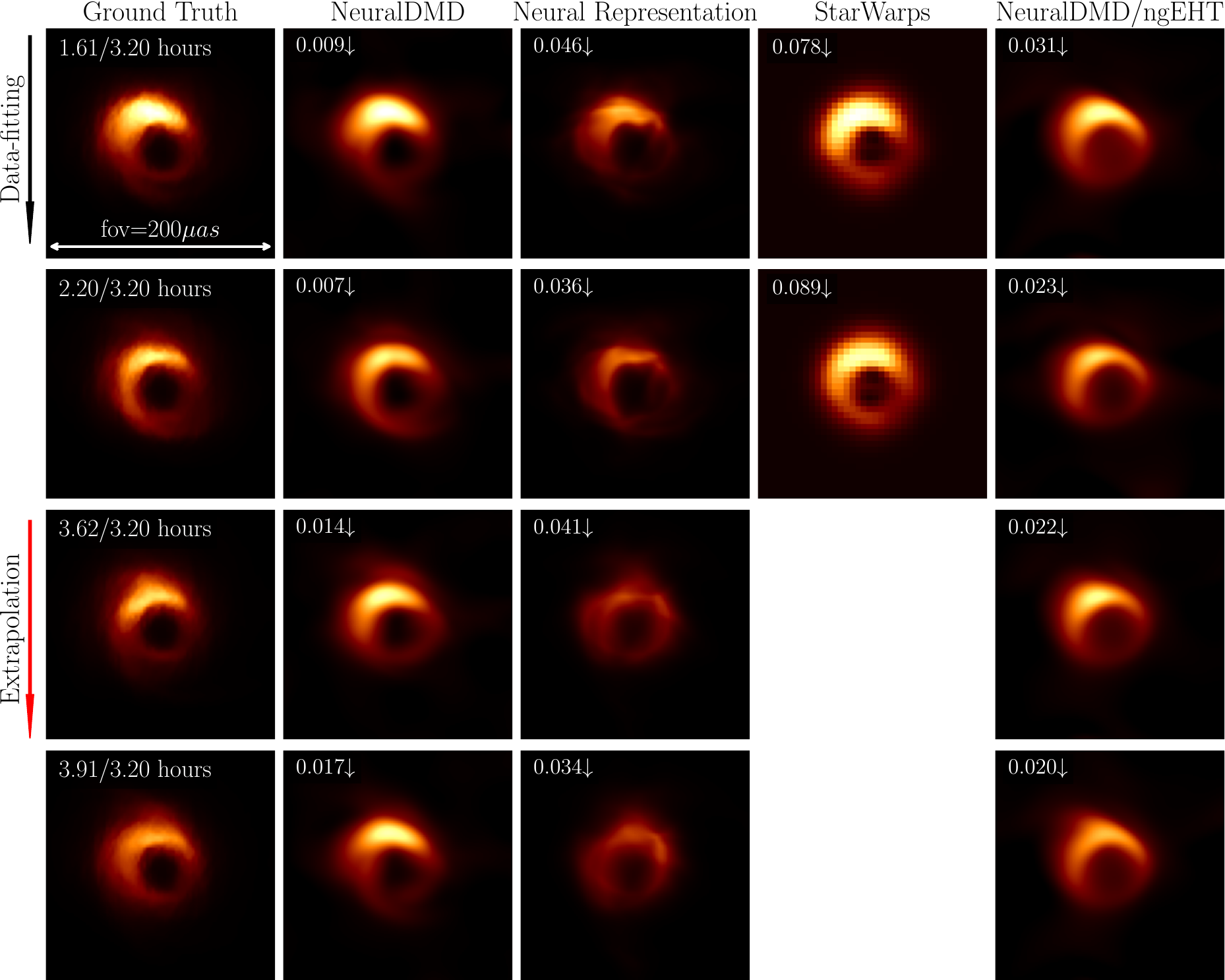}
    \caption{Reconstruction results on a GRMHD with interferometric observations synthesized for the \ngehtp array. From left to right, the columns show the ground truth, \neuralDMD, neural representation, \sw, \neuralDMD reconstruction from \ngeht array. In each row, the first two frames are reconstructed from the observations, i.e., constrained by data for fitting the model. The second two frames are extrapolation results obtained by advancing the recovered dynamical (or neural) model forward in time to unobserved frames. The value at the top left corner of each frame shows the normalized L2-error of that reconstruction as compared to the ground truth (lower is better).}
    \label{fig: GRMHD2 tests ngEHT_plus}
\end{figure*}

\begin{figure}
    \centering
    \includegraphics[width=1.0\linewidth]{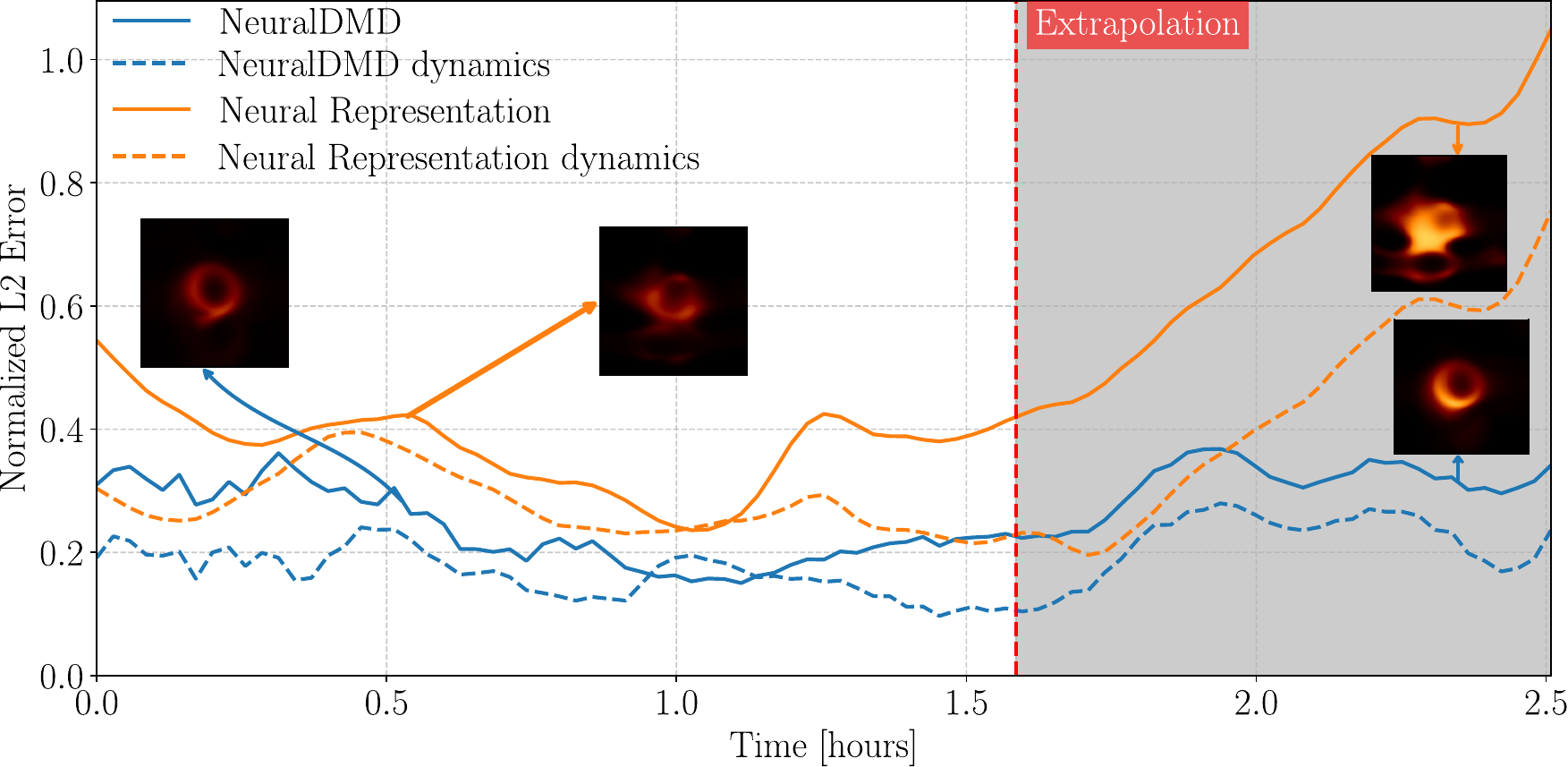}
    \caption{Reconstruction error for \neuralDMD (blue) and the neural representation (orange) for \ngeht coverage. Solid curves give total error, dashed curves the dynamics‑only error (mean image removed), and reconstructed frames are shown at selected points. Data-fitting relies on observations from a $1.6$ hour window; the grey band marks the extrapolation time window.  The neural representation error exhibits rapidly increasing error beyond the observational window, whereas \neuralDMD maintains stable reconstructions -- preserving both the average structure and dynamic behavior -- over extended time horizons.}
    \label{fig: extra_err}
\end{figure}

In the main text, we focus on \neuralDMD reconstructions of \ehtobs, where the baseline neural representation is pretrained on a uniform disk and \neuralDMD’s spatial modes are initialized with Zernike polynomials. Here, we extend the comparison to denser arrays—first \ngeht and then an augmented configuration \ngehtp—to assess how \neuralDMD performs as $(u,v)$ coverage improves. For \ngeht and \ngehtp, both the baseline neural representation and \neuralDMD are initialized randomly rather than from a pretrained disk.
The planned upgrades -- namely, the next generation EHT (\ngeht) \cite{ngEHT} and the space-borne Black Hole Explorer \cite{BHExplorer} -- aim to significantly improve baseline coverage. This increased sampling density would reduce the need for time averaging, which is typically used to compensate for sparse data but can blur the horizon-scale variability in dynamic sources~\cite{EHT_SgrA_III_2023}.

\autoref{fig: GRMHD2 tests} and \autoref{fig: GRMHD2 tests ngEHT_plus} show the reconstruction results of \neuralDMD, neural representation, \sw, as well as \neuralDMD with \ngeht and \ngehtp coverage, respectively.  It also compares the performance of these methods for extrapolating into the future. \autoref{fig: extra_err} shows that even with the improved \ngeht coverage, neural representation extrapolation error quickly blows up, where as \neuralDMD's reconstructions remain stable. 

\section{Additional Dynamical Analysis}
\label{sec: Analysis}

\begin{figure*}[t]
    \centering
    \includegraphics[width=1.0\linewidth]{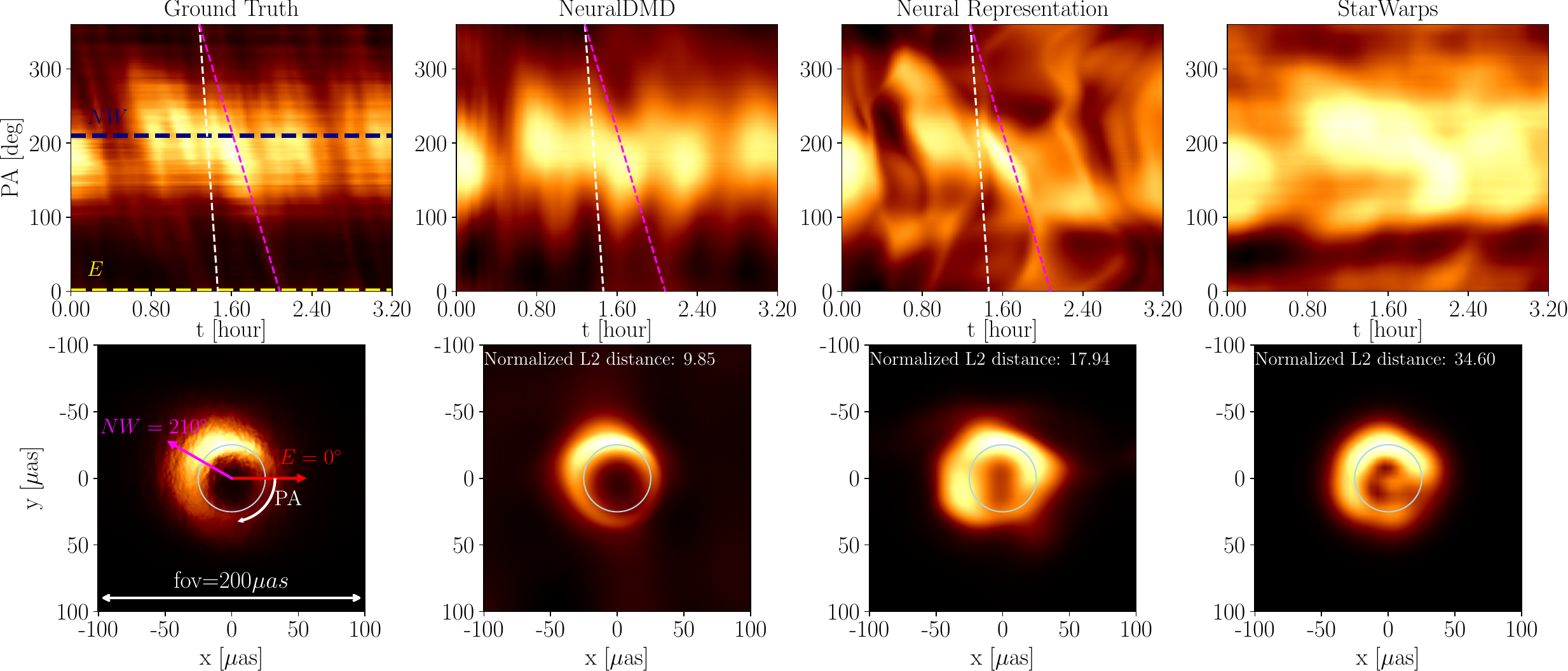}
    \caption{\ehtobs coverage cylinder plots. The top row panels show the cylinder plots along the ring at the radius of peak emission. The second row shows the average frames where the selected ring used for the cylinder plot is depicted. The position angle (PA) is measured clockwise from the direction of East ($E$) and NorthWest ($NW$), is highlighted in both for orientation. The white dashed line shows an analytical counter-clockwise Keplerian angular velocity, while the magenta denotes a sub-Keplerian (slower) angular velocity extracted from an image feature. Note that we do not show these lines for neural representation, and \sw as the poor reconstruction quality prohibits precise angular velocity measurements.}
    \label{fig: cylinder_eht2017}
\end{figure*}

\begin{figure*}[t]
    \centering
    \includegraphics[width=1.0\linewidth]{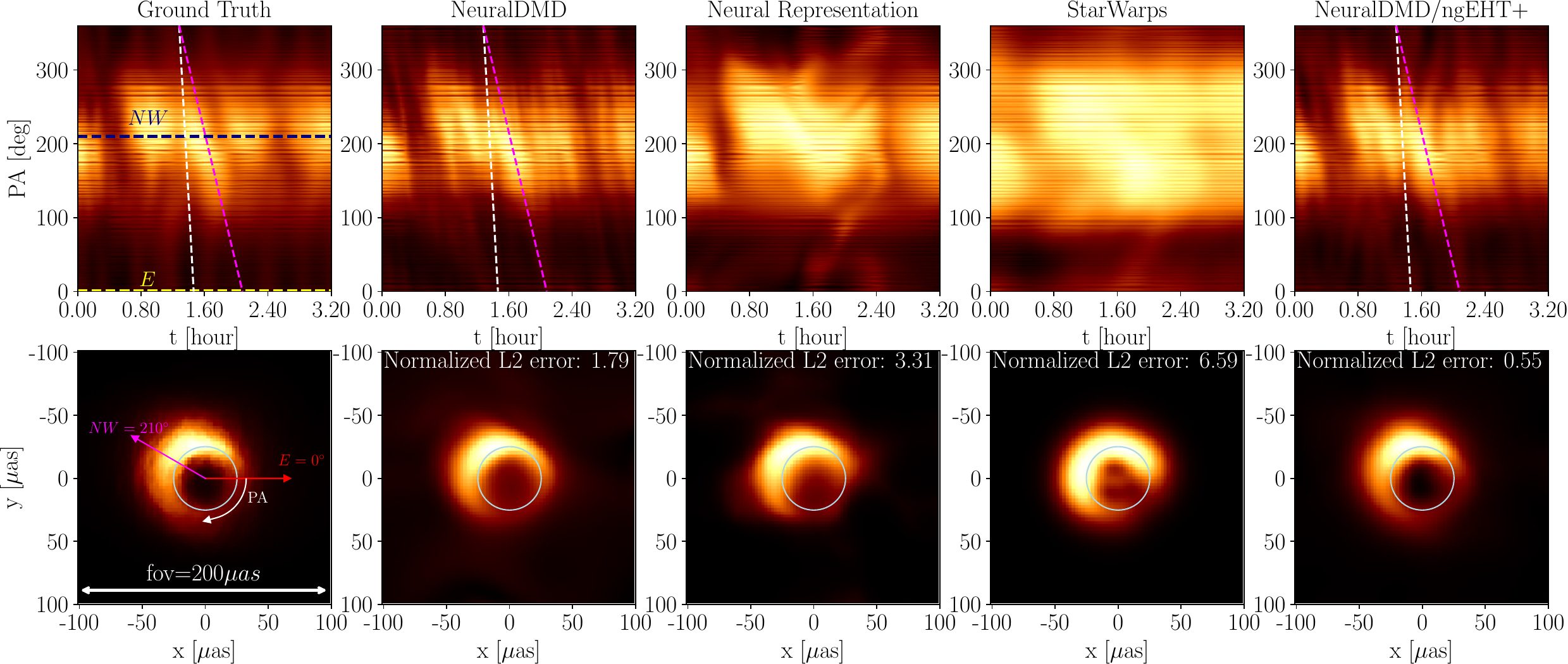}
    \caption{\ngeht coverage cylinder plots. The top row panels show the cylinder plots along the ring at the radius of peak emission. The second row shows the average frames where the selected ring used for the cylinder plot is depicted. The position angle (PA) is measured clockwise from the direction of East ($E$) and NorthWest ($NW$), is highlighted in both for orientation. The white dashed line shows an analytical counter-clockwise Keplerian angular velocity, while the magenta denotes a sub-Keplerian (slower) angular velocity extracted from an image feature. Note that we do not show these lines for neural representation, and \sw as the poor reconstruction quality prohibits precise angular velocity measurements.}
    \label{fig: cylinder_plots}
\end{figure*}

\begin{figure*}[t]
    \centering
    \includegraphics[width=1.0\linewidth]{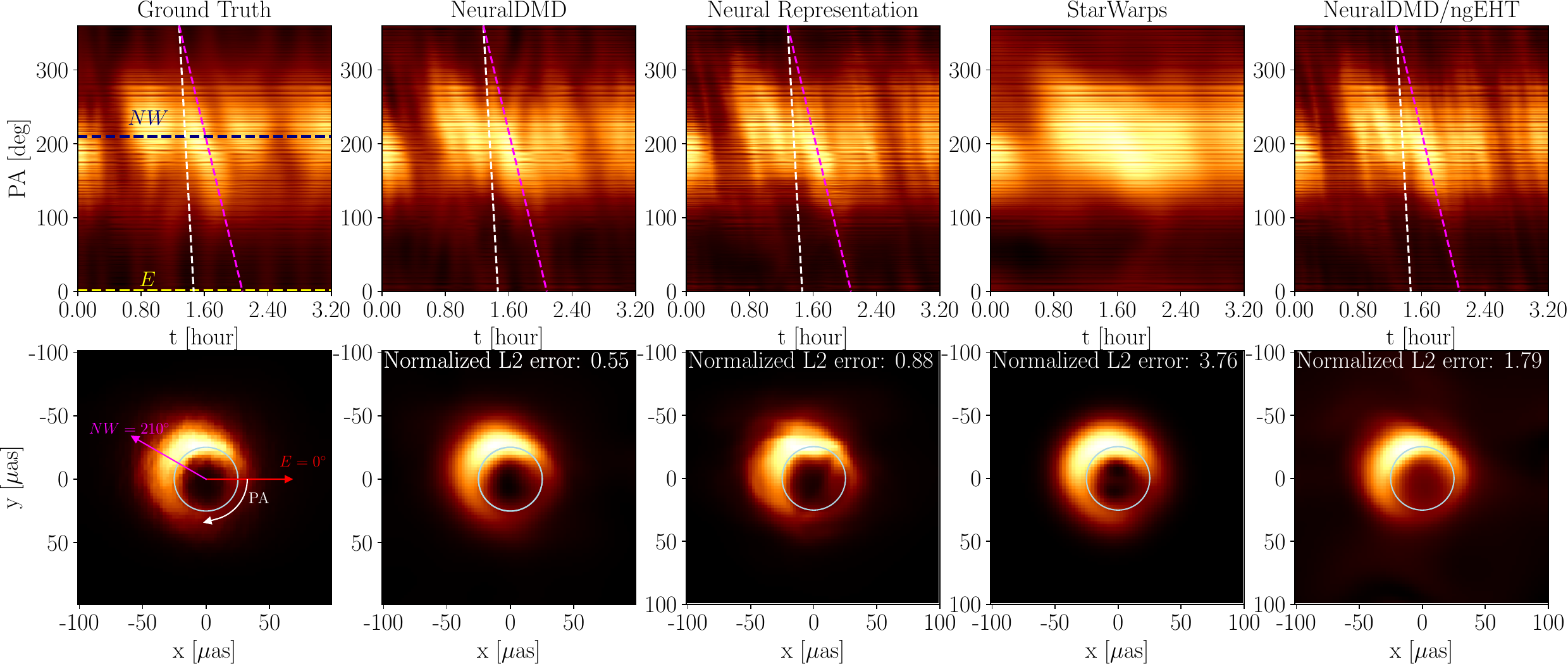}
    \caption{\ngehtp coverage cylinder plots. First row shows the cylinder plots at the radius of the maximum emission. The second row shows the average frames where the selected ring is depicted. The white dashed line shows Keplerian angular velocity, while the magenta denotes a sub-Keplerian angular velocity of the point of maximum emission for the ground truth due to its smaller slope. We do not show these lines for \sw as the reconstructions are not fine enough to find the anuglar velocity of these features. The reconstructions are from \ngehtp coverage unless otherwise stated.}
    \label{fig: cylinder_plots_ngEHT_plus}
\end{figure*}

\begin{figure*}[t]
    \centering
    \includegraphics[width=1.0\linewidth]{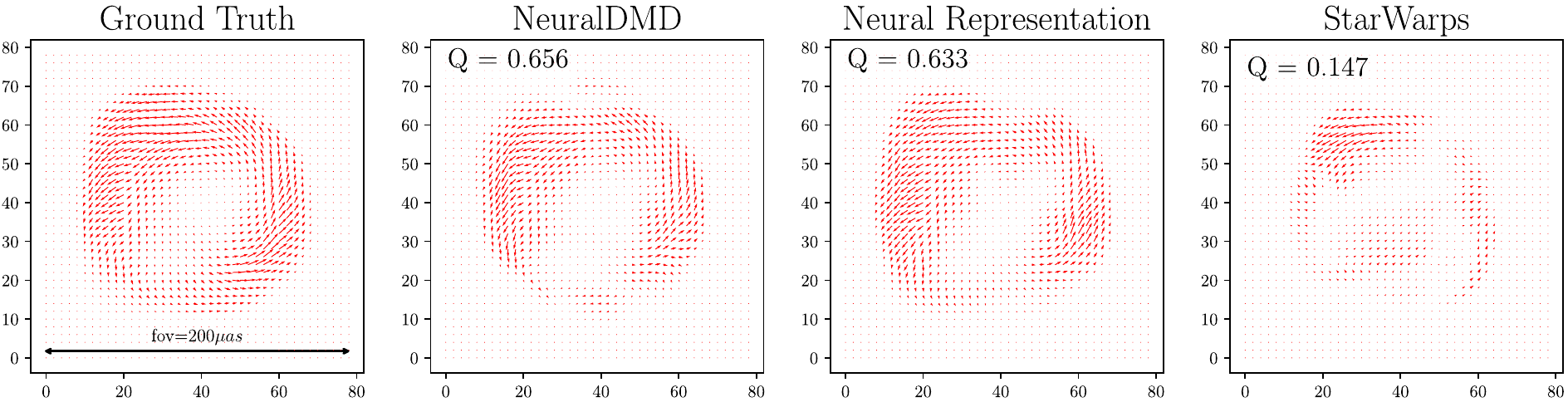}
    \caption{Mean optical flows of ground truth, and \ngehtp coverage reconstruction results of \neuralDMD, neural representation, StarWarps, and \ngeht coverage reconstruction result of \neuralDMD. The value $Q$ shows the cosine similarity (higher is better) of the reconstructions with the ground truth. Note that $Q$ does not capture the noise reconstruction; hence, even though $Q$ for neural representation is close to \neuralDMD, it hallucinates dynamics, resulting in a poor reconstruction quality.}
    \label{fig: optical_flows_ngEHT_plus}
\end{figure*}

In the previous sections and the main text, we highlighted the ability of \neuralDMD to recover image frames (and even extrapolate into the future), reconstructing spatial structures that are quantitatively and qualitatively close to the ground-truth GRMHD simulation. In this section, we further analyze the quality of the recovered dynamics with two types of analyses.

\textbf{Cylinder plot:} for a black hole, most of the visible emission is expected to come from the region around the photon sphere. For Sagittarius A*, this region, when projected onto the image plane, is expected to be at $r\simeq 25 \mu as$. We sample the image plane azimuthally at this radius and ``unwrap'' the pixel values along the ring as a function of time. This coordinate transformation produces the cylinder plots illustrated in Fig.~\ref{fig: cylinder_plots}. In the cylinder plot, a counter-clockwise bright feature at a constant orbital velocity will appear as a bright streak (line) with a negative slope. Precise reconstructions of cylinder plots would enable estimating the underlying physical origins of rotating features (e.g., shock waves vs. hotspots)~\cite{conroy2023rotation} enabling studying these not well understood plasma dynamics from observations (rather than simulations). In Fig.~\ref{fig: cylinder_plots}, we demonstrate that \neuralDMD is able to capture the velocity of these dynamic structures, outperforming both baselines where the streaking lines are lost. The well-defined streaking lines in the cylinder plot produced by \neuralDMD enable clearer and more accurate angular velocity estimation. For context, we overlay the Keplerian angular velocity curve, highlighting deviations that indicate physical processes beyond pure gravitational rotation. 

\textbf{Optical Flow:} 
While the cylinder analysis focuses on dynamics along the brightest region of the ring, in principle optical flow enables recovering arbitrary non-rotational velocity fields across the entire image plane. Figure \ref{fig: optical_flows_ngEHT_plus} shows the \emph{mean optical flow}, which quantifies the average apparent motion of pixel intensities between successive image frames. In the main text (Sec.~\ref{subsec:fourier_domain}), we compare the mean optical flow of the ground-truth GRMHD to \neuralDMD and two baselines neural representation and \sw to further assess and quantify performance. The mean optical flow of the \ngeht coverage is included in the main text. With \ngeht observation coverage, distinct performance trends emerge: \sw avoids artificial dynamics due to the temporal regularization; however, it misses significant dynamic features for the same reason. The neural representation tends to overfit sparse, noisy data, introducing spurious dynamics that deviate from the ground truth. In contrast, \neuralDMD is able to more accurately reconstruct the underlying flow. Furthermore, we explore the improved observational coverage from \ngehtp (figure \ref{fig: optical_flows_ngEHT_plus}) and demonstrate that it significantly enhances reconstruction quality, yielding mean optical flows closer to the ground truth. The latter result suggests that considering the specifics of the reconstruction algorithm might be crucial when planning the expansion of the EHT array.

In summary, \autoref{fig: cylinder_eht2017}, \autoref{fig: cylinder_plots}, and \autoref{fig: cylinder_plots_ngEHT_plus} demonstrate how different methods perform in predicting the angular velocity of one of the brightest features with \ehtobs, \ngeht, and \ngehtp coverage, respectively. Furthermore, \autoref{fig: optical_flows_ngEHT_plus} depicts the mean optical flow of all the methods for \ngehtp, and provides a comparison of the captured dynamics. The mean optical flow for \ngeht is included in the last column of \autoref{fig: GRMHD2 tests}.

\section{Hyperparameters and Training Details}
Here, we briefly discuss the hyperparameters we use for training \neuralDMD. We multiply $\Omega t$ with a constant value of order $10^2$ to bring the $\Omega$ values to a physical range, as well as increase the values of the gradients of loss function with respect to the $\Omega$'s. In all experiments, the initial learning rate is $1e-4$, and we use a learning rate scheduler that divides the learning rate by a factor of $2$ if the loss plateaus for $1000$ epochs. We trained our models for about $12,000$ epochs. Every experiment took about an hour to fully train; however, a mere $5-20$ minutes on one GeForce RTX 4070 GPU would still lead to high-quality reconstructions. However, on an RTX 4070 GPU, due to memory issues, we are only able to train with very small batch sizes. Training with one NVIDIA H200 Tensor Core GPU allows us to use arbitrarily high batch sizes, and reduces the training time to about $5$ minutes for $12,000$ epochs.
\section{Robustness Experiments}
\label{sec: robustness experiments}
\begin{figure*}
    \centering
    \includegraphics[width=0.85\linewidth]{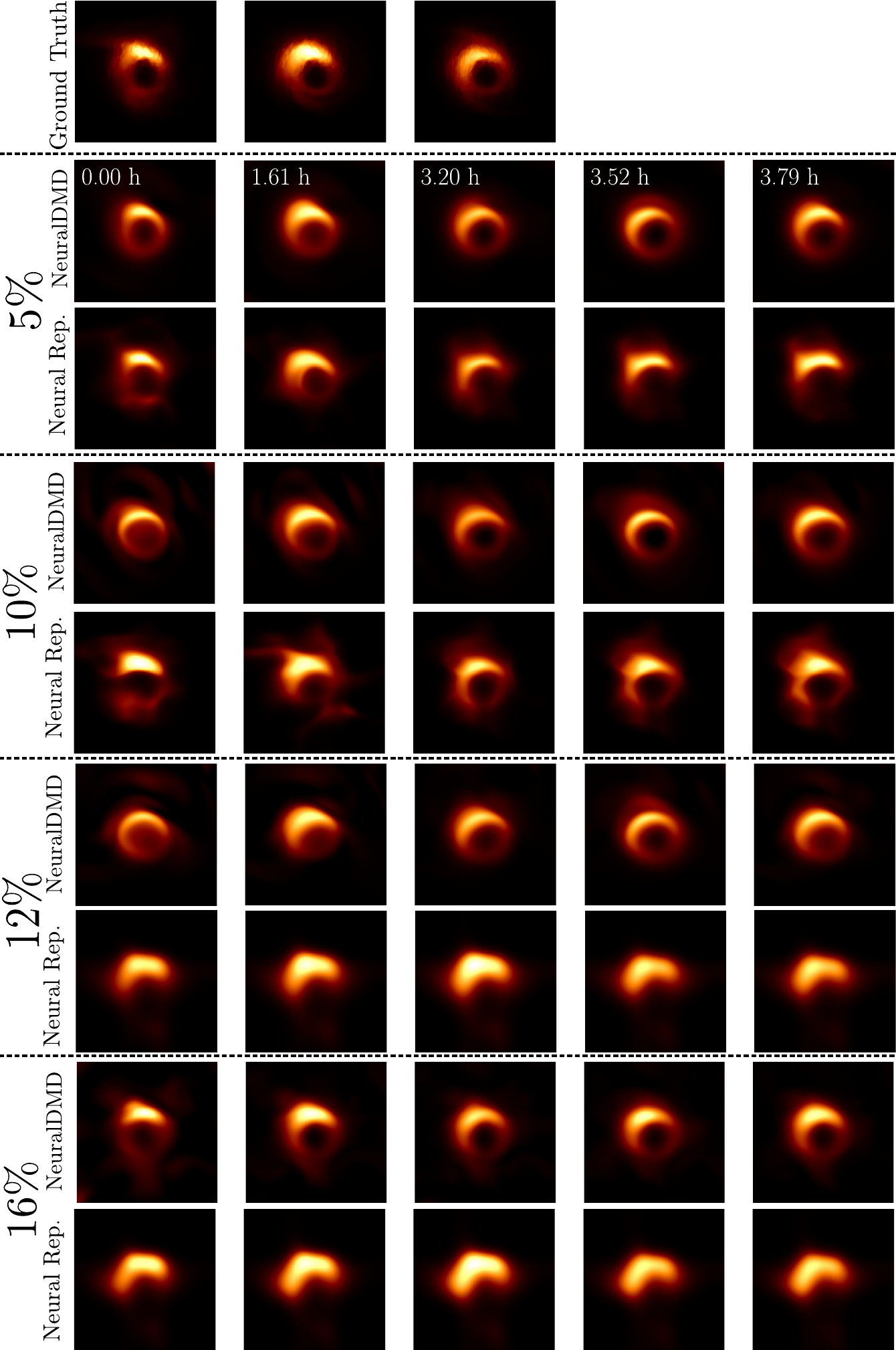}
    \caption{\neuralDMD and neural representation (Neural Rep.) imaging performance under different amounts of noise. \neuralDMD consistently outperforms neural representation. The percentage shows the amount of fractional noise added to the data. The two right-most columns are extrapolation after the data-fitting window.}
    \label{fig: noise exps}
\end{figure*}

\begin{figure*}
    \centering
    \includegraphics[width=0.85\linewidth]{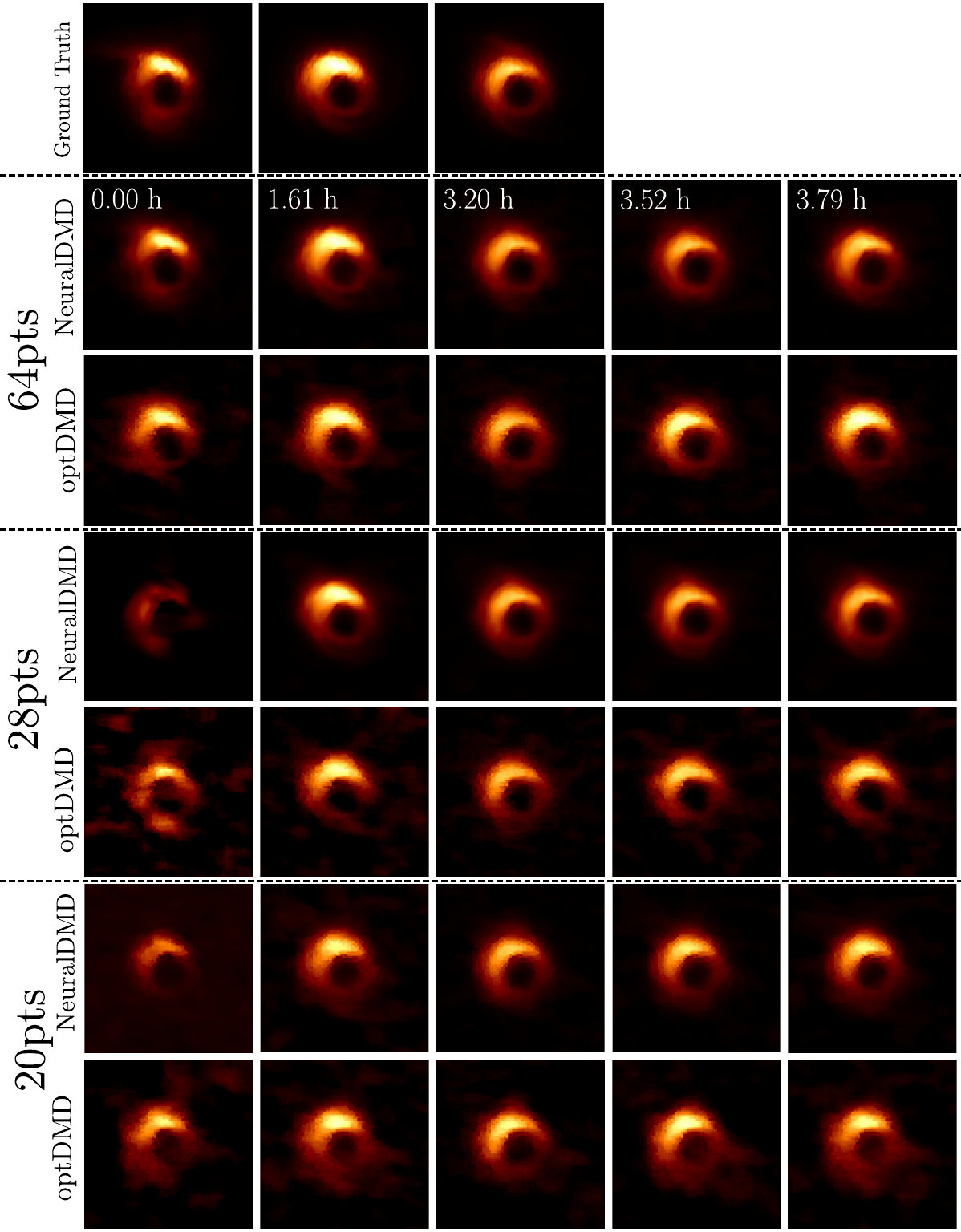}
    \caption{\neuralDMD and \optdmd imaging performance under different sparsity settings. \neuralDMD outperforms TV-regularized \optdmd (while there are some numerical issues for some of the first frames). The number in the format $num$pts is the number of Fourier frequencies used in data fitting for each frame. The two right-most columns are extrapolation after the data-fitting window.}
    \label{fig: sparsity exps}
\end{figure*}

\begin{figure*}
    \centering
    \includegraphics[width=0.9\linewidth]{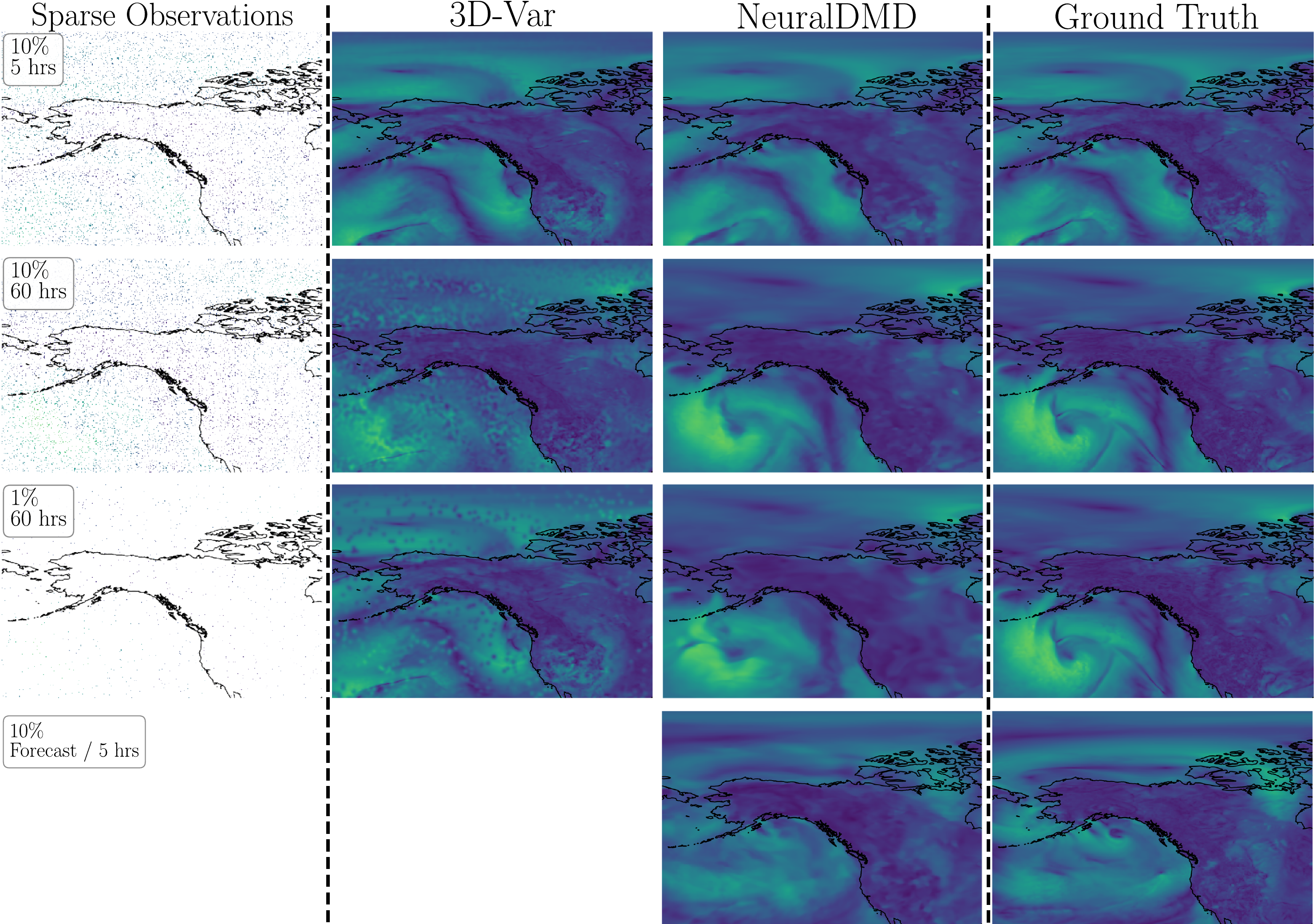}
    \caption{A comparison of \neuralDMD with the data assimilation baseline \varbaseline. The results are shown on sparsely sampled ERA5 datasets of wind velocity magnitudes. [\textbf{First row}]: Results for $5$ hours from the initial frame [\textbf{Second row}] $60$ hours from the initial frame with $10\%$ of data given as sparse observations, [\textbf{Third Row}] $60$ hours from the initial frame with $1\%$ of the data given. [\textbf{Fourth row}] demonstrates \neuralDMD’s ability to extrapolate by forecasting 5 hours beyond the observation window. From left to right: sparsely sampled observations, \varbaseline, and \neuralDMD reconstructions, and ground truth.}
    \label{fig: weather baseline}
\end{figure*}

In the main text, we show \neuralDMD outperforms \optdmd and neural representation under the same noise and sparsity levels. Here, we expand the comparisons under varying noise and sparsity levels. Figure~\ref{fig: noise exps} demonstrates the robustness of \neuralDMD under varying noise levels as compared to neural representations, while figure~\ref{fig: sparsity exps} compares \neuralDMD under varying sparsity levels with \optdmd.
Moreover, in \autoref{fig: weather baseline}, for the image-domain sparsity experiments of weather dynamics, we test \neuralDMD under two levels of sparsity: sampling only $10\%$ or $1\%$ of the grid points. In both cases, we compare reconstructions with our version of \varbaseline.

\section{Edge-on GRMHD}
\label{sec: edge-on grmhd}
\begin{figure*}
    \centering
    \includegraphics[width=0.85\linewidth]{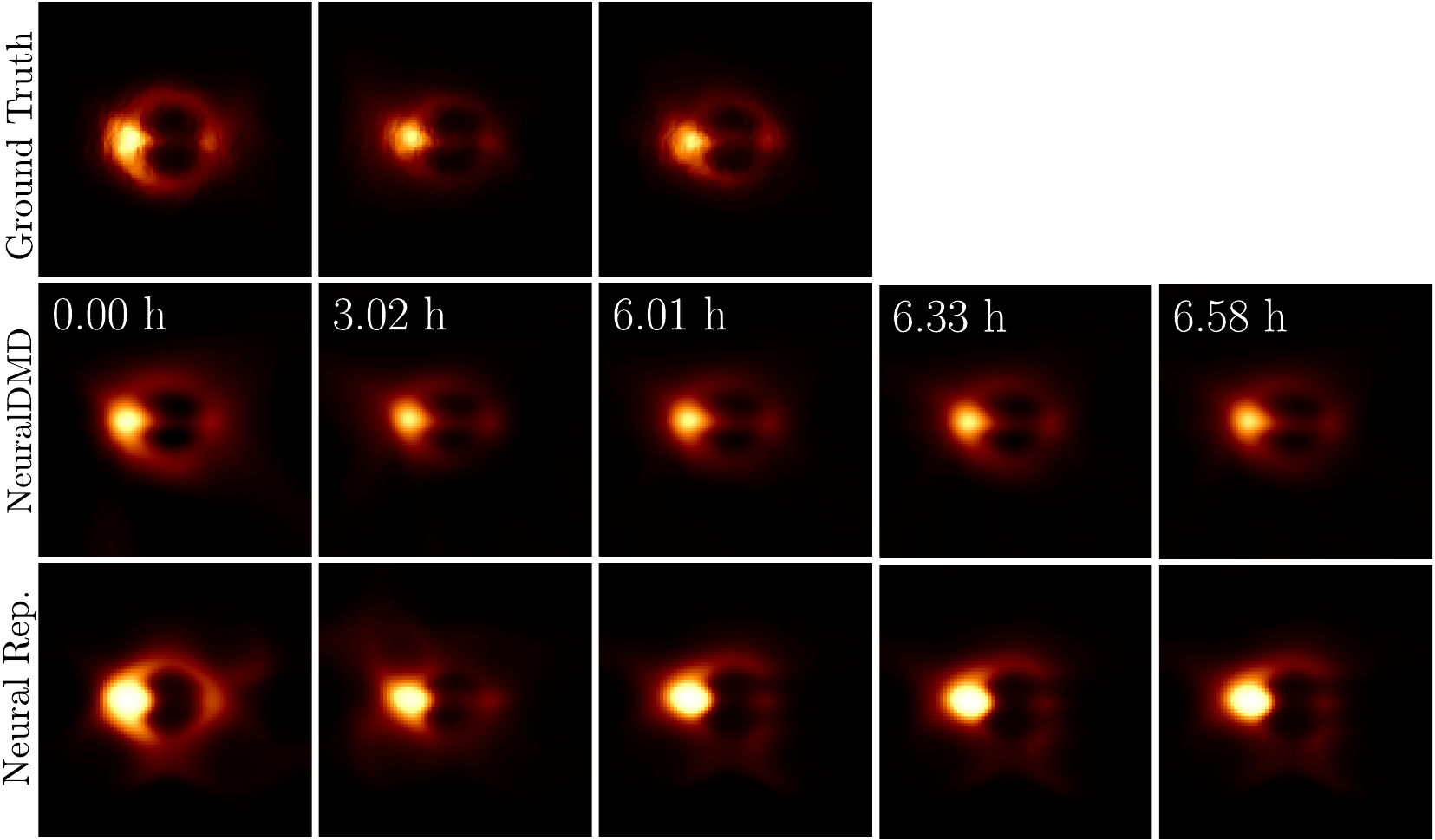}
    \caption{\neuralDMD and neural representation (Neural Rep.) imaging performance on an edge-on GRMHD simulation. This test is more challenging than a face-on GRMHD since the rotation dynamics of the accretion moves in and out of the image plane as the line of observation is almost parallel to the image plane. Whereas, in the face-on case, the majority of the dynamics is well-captured within the image plane.}
    \label{fig: edge-on}
\end{figure*}

So far, we limited our discussion to face-on GRMHD simulations, where most of the dynamics is well-captured on the image plane. A more challenging simulation is called an edge-on GRMHD, which is characterized by the movement of the accretion of the black hole in and out of the image plane. We show that \neuralDMD is still able to successfully model dynamics for such simulations under \ngeht coverage and outperform a pure neural representation baseline (figure \ref{fig: edge-on}).
\section{Orbiting Hot Spot}
\begin{figure*}[t]
    \centering
    \includegraphics[width=1.0\linewidth]{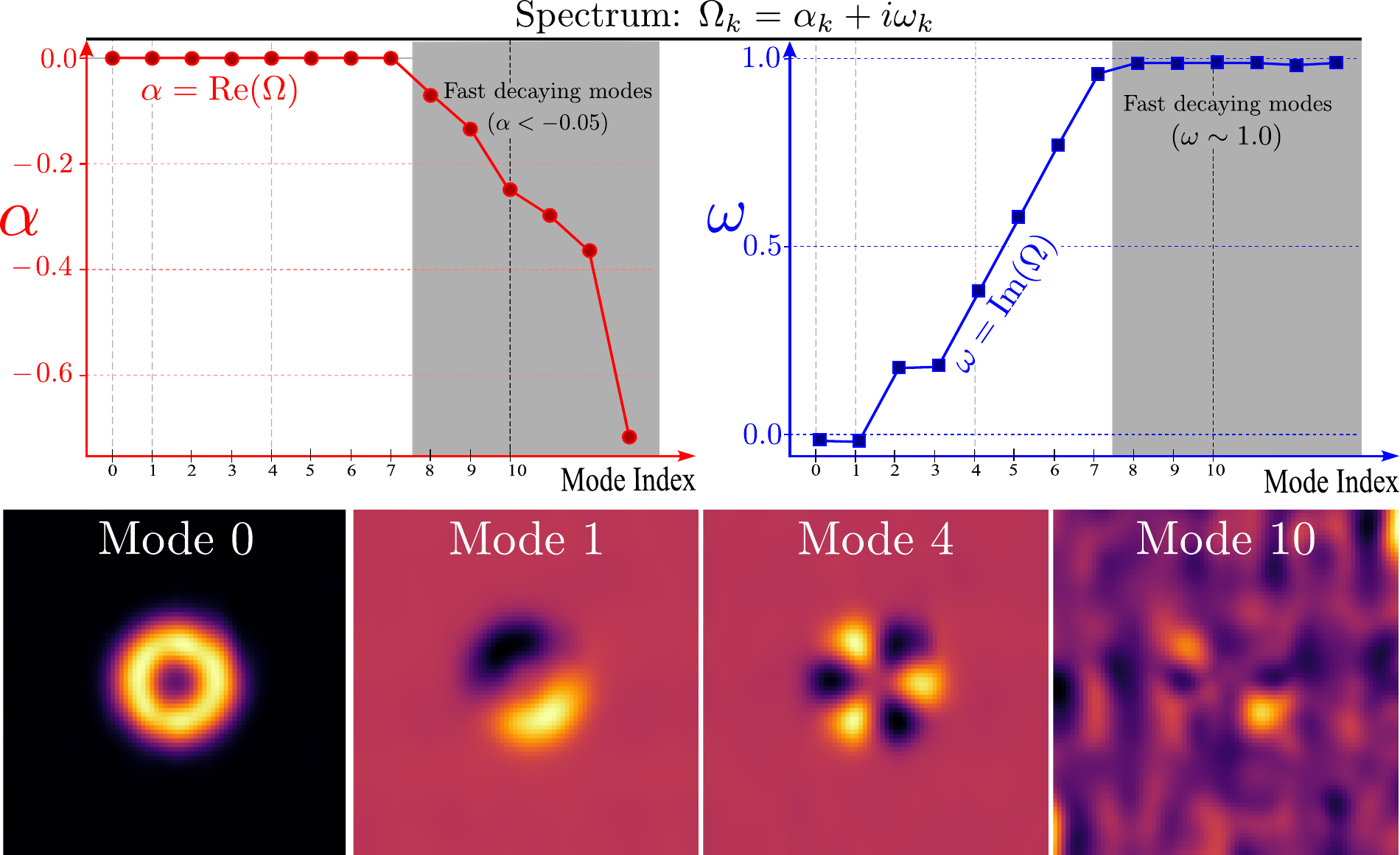}
        \caption{Orbiting hot spot modes (real part) and eigenvalues. The top panel shows both the real (red) and imaginary (blue) parts of the eigenvalues. The imaginary part of the eigenvalues, $\omega = \text{Im}(\Omega)$, indicates the oscillatory behavior and the real part, $\alpha = \text{Re}(\Omega)$, indicates the decay rate. The gray-shaded region highlights higher-order modes with fast decay ($\alpha < -0.05$) which do not impact the reconstruction. The bottom panel showcases four of the recovered \neuralDMD modes, illustrating the multi-scale nature of the modal decomposition.}
    \label{fig: hs omegas_modes}
\end{figure*}
Occasionally, the plasma dynamics around Sagittarius A*, the black hole at the center of our galaxy, emit strong energetic flares visible in X-ray~\cite{neilsen2013chandra}, infrared~\cite{fazio2018multiwavelength}, and radio wavelengths~\cite{Wielgus2022lc, levis2024orbital}. One explanation for these flares is compact (localized) bright emission regions, referred to as ``hot-spots''~\cite{broderick2005imaging}, which form and orbit the black hole. Accurate dynamical imaging of these transient features could enable probing the strong gravitational field and spacetime geometry near the event horizon~\cite{tiede2020spacetime}.

As a simple test case, we create synthetic observations of a Gaussian hot spot model. We generate a hot spot at a distance of $25\mu as$ from the center, orbiting counterclockwise with an angular velocity of approximately $9.8 \text{rad}/s$. The total flux intensity of the hot spot is normalized to $1$Jy, with intensity decreasing from the center of the hot spot following a standard deviation of $10 \mu as$. We recover the modes $w_j$ and spectrum $\Omega_k$ using the \neuralDMD minimization described in Sec.~\ref{sec: Methods}. \neuralDMD manages to perfectly reconstruct the motion of the hot spot from the sparse observations and it can perfectly extrapolate the motion of the hot spot for arbitrary times into the future.

Figure \ref{fig: hs omegas_modes} illustrates the neural modes $w_j$, and the complex spectrum $\Omega_j$, recovered from the sparse \ngehtp observations of the synthetic hot spot. The recovered modes provide a multi-representation where $w_j$ captures the time average of the orbiting hot spot (seen as a ring). Higher-order modes capture a more nuanced spatial spectrum and dynamics. By constraining $\alpha \leq 0$ \neuralDMD is able to recover all modes simultaneously while ensuring the stability of the recovered dynamics. We a priori set the number of modes and discard high-order fast decaying modes (very negative $\alpha$ values) in a post-processing step (see Fig.~\ref{fig: hs omegas_modes}). Interestingly, the frequency of the first dynamic mode (mode $1$; the double crescent shape) is exactly equal to the angular frequency of the orbiting hot spot, while the higher order modes are multiples of the orbital frequency. Therefore, at least for the case of one orbiting feature, the modes and their frequencies is completely interpretable.

\section{Fourier Frequency Coverage Pattern}
\label{sec: coord tables and patterns}
\begin{figure*}[t]
    \centering
    \includegraphics[width=1.0\linewidth]{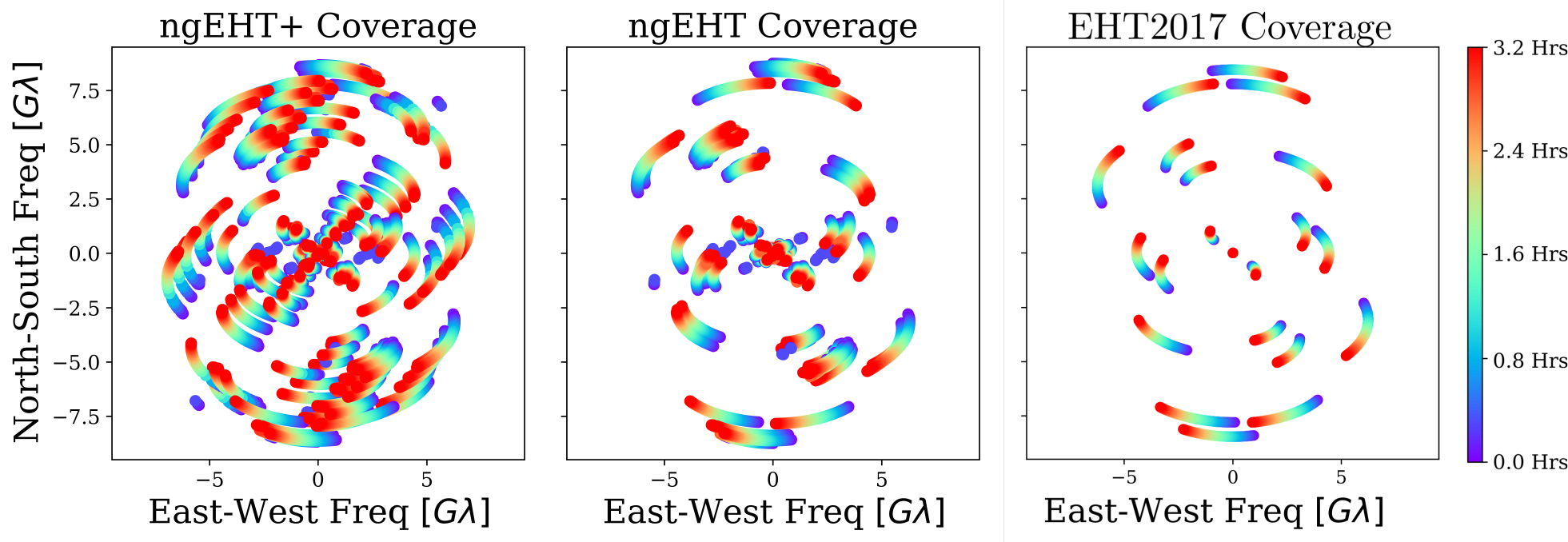}
    \caption{Frequency coverage of \ehtobs, \ngeht, and the enhanced \ngehtp. The scatter plots overlay the observed frequencies collected during the entire time window of 3.2 hours, color-coded by time. Note that the uv sampling smoothly changes as the Earth rotates throughout the acquisition period.}
    \label{fig: coverage arrays}
\end{figure*}

\begin{table*}[t]
  \centering
  \setlength{\tabcolsep}{4pt}
  \renewcommand{\arraystretch}{1.0}
  \footnotesize

  \begin{minipage}[t]{0.47\linewidth}
    \centering

    {\bfseries Selected Stations for \ehtobs\par}
    \vspace{0.3em}
    \resizebox{\linewidth}{!}{
    \begin{tabular}{lrrr}
      \toprule
      \textbf{Station} & \textbf{x (m)} & \textbf{y (m)} & \textbf{z (m)}\\
      \midrule
      PV   & 5,088,967.900 &   -301,681.600 &  3,825,015.800 \\
      SMT  & -1,828,796.200 & -5,054,406.800 & 3,427,865.200 \\
      SMA  & -5,464,523.400 & -2,493,147.080 & 2,150,611.750 \\
      LMT  &   -768,713.964 & -5,988,541.798 & 2,063,275.947 \\
      ALMA & 2,225,061.164 & -5,440,057.370 & -2,481,681.150 \\
      APEX & 2,225,039.530 & -5,441,197.630 & -2,479,303.360 \\
      JCMT & -5,464,584.680 & -2,493,001.170 & 2,150,653.980 \\
      SPT  &        0.010  &        0.010   & -6,359,609.700 \\
      \bottomrule
    \end{tabular}
    }

    \vspace{0.9em}

    {\bfseries Selected Stations for \ngeht\par}
    \vspace{0.3em}
    \resizebox{\linewidth}{!}{
    \begin{tabular}{lrrr}
      \toprule
      \textbf{Station} & \textbf{x (m)} & \textbf{y (m)} & \textbf{z (m)}\\
      \midrule
      ALMA & 2,225,061.164 & -5,440,057.370 & -2,481,681.150 \\
      APEX & 2,225,039.530 & -5,441,197.630 & -2,479,303.360 \\
      BAJA & -2,352,576.000 & -4,940,331.000 & 3,271,508.000 \\
      CNI  & 5,311,000.000 & -1,725,000.000 & 3,075,000.000 \\
      GAM  & 5,648,000.000 &  1,619,000.000 & -2,477,000.000 \\
      GLT  & 1,500,692.000 & -1,191,735.000 & 6,066,409.000 \\
      HAY  & 1,521,000.000 & -4,417,000.000 & 4,327,000.000 \\
      JCMT & -5,464,584.680 & -2,493,001.170 & 2,150,653.980 \\
      KP   & -1,995,678.840 & -5,037,317.697 & 3,357,328.025 \\
      LLA  & 2,325,338.000 & -5,341,460.000 & -2,599,661.000 \\
      LMT  &  -768,713.964 & -5,988,541.798 & 2,063,275.947 \\
      OVRO & -2,409,598.000 & -4,478,348.000 & 3,838,607.000 \\
      SMA  & -5,464,523.400 & -2,493,147.080 & 2,150,611.750 \\
      SMT  & -1,828,796.200 & -5,054,406.800 & 3,427,865.200 \\
      SPT  &        0.010  &        0.010   & -6,359,609.700 \\
      \bottomrule
    \end{tabular}
    }
  \end{minipage}
  \hfill
  \begin{minipage}[t]{0.50\linewidth}
    \centering

    {\bfseries Selected Stations for \ngehtp\par}
    \vspace{0.3em}
    \resizebox{\linewidth}{!}{
    \begin{tabular}{lrrr}
      \toprule
      \textbf{Station} & \textbf{x (m)} & \textbf{y (m)} & \textbf{z (m)}\\
      \midrule
      ALMA & 2,225,061.164 & -5,440,057.370 & -2,481,681.150 \\
      APEX & 2,225,039.530 & -5,441,197.630 & -2,479,303.360 \\
      GLT  & 1,500,692.000 & -1,191,735.000 & 6,066,409.000 \\
      JCMT & -5,464,584.680 & -2,493,001.170 & 2,150,653.980 \\
      KP   & -1,995,678.840 & -5,037,317.697 & 3,357,328.025 \\
      LMT  &  -768,713.964 & -5,988,541.798 & 2,063,275.947 \\
      SMA  & -5,464,523.400 & -2,493,147.080 & 2,150,611.750 \\
      SMT  & -1,828,796.200 & -5,054,406.800 & 3,427,865.200 \\
      SPT  &        0.010  &        0.010   & -6,359,609.700 \\
      BAJA & -2,352,576.000 & -4,940,331.000 & 3,271,508.000 \\
      BAR  & -2,363,000.000 & -4,445,000.000 & 3,907,000.000 \\
      CAT  & 1,569,000.000 & -4,559,000.000 & -4,163,000.000 \\
      CNI  & 5,311,000.000 & -1,725,000.000 & 3,075,000.000 \\
      GAM  & 5,648,000.000 &  1,619,000.000 & -2,477,000.000 \\
      GARS & 1,538,000.000 & -2,462,000.000 & -5,659,000.000 \\
      HAY  & 1,521,000.000 & -4,417,000.000 & 4,327,000.000 \\
      NZ   & -4,540,000.000 &   719,000.000 & -4,409,000.000 \\
      OVRO & -2,409,598.000 & -4,478,348.000 & 3,838,607.000 \\
      SGO  & 1,832,000.000 & -5,034,000.000 & -3,455,000.000 \\
      CAS  & 1,373,000.000 & -3,399,000.000 & -5,201,000.000 \\
      LLA  & 2,325,338.000 & -5,341,460.000 & -2,599,661.000 \\
      PIKE & -1,285,000.000 & -4,797,000.000 & 3,994,000.000 \\
      PV   & 5,088,967.900 &   -301,681.600 & 3,825,015.800 \\
      \bottomrule
    \end{tabular}
    }
  \end{minipage}

  \caption{Selected station coordinates for \ehtobs, \ngeht, and \ngehtp.}
  \label{tab:selected_stations_all}
\end{table*}
The main text relied on the sparse visibility coverage of the \ehtobs array. In the previous sections, we repeated the experiments with an expanded configuration, namely \ngeht, and \ngehtp which augments \ngeht with additional stations, allowing us to assess how denser baseline coverage enhances our ability to recover source dynamics. Table~\ref{tab:selected_stations_all} shows the stations used in \ehtobs, \ngeht, and \ngehtp array coverages used in the main text and here in Supp., respectively, along with their Cartesian coordinates with the center of Earth as the origin. Note that the addition of only a few more stations, as showcased in \ngeht and even more prominently \ngehtp, leads to much higher quality reconstructions with \neuralDMD. The Fourier sampling pattern for all three cases are shown in Fig.~\ref{fig: coverage arrays}.

\clearpage
\section{Gain Marginalization}
\label{sec: gain marginalization}
\subsection{Background and Definitions}
The model visibilities for stations $i$ and $j$, $V_{ij} (\theta)$ are related to observed dirty observations via
\begin{equation}
    \begin{aligned}
    V_{ij}^{\text{obs}} &= g_i g_j^* V_{ij}(\theta) + n_{ij}, \,\, n_{ij} \sim \mathcal{N}(0, \sigma_{ij}^2),\\
    g_k &\equiv \exp{(a_k + i \phi_k)}, \,\,\, a_k, \phi_k \in \mathbb{R},
    \end{aligned}
\end{equation}
where $a_k$ and $\phi_k$ respectively show the amplitude and phase contributions of the complex gain on station $k$, $g_k$. $n_{ij}$ captures the thermal noise over the baseline measurements which is due to telescope sensitivities. 

Assuming $M$ baselines have been used during observations and $N$ stations, we can stack all observations and formulate a unified vector as
\begin{equation}
    \begin{aligned}
        y &= \begin{bmatrix}
        \text{Re}(V^{\text{obs}})\\
        \text{Im}(V^{\text{obs}})
    \end{bmatrix},
    \,\, y_0 (\theta) = \begin{bmatrix}
        \text{Re}(V)\\
        \text{Im}(V)
    \end{bmatrix},\\
    \text{residual: } r &\equiv y - y_0 (\theta).
    \end{aligned}
\end{equation}

Linearization about the current gains ($a_k = a_0, \,\, \phi_k = \phi_0$) gives
\begin{equation}
    \begin{aligned}
        \Delta V_{ij} &\approx (a_i + a_j + i (\phi_i - \phi_j)) V_{ij},\\
        V_{ij} &\equiv g_{0, i} g_{0, j}^*  V_{ij}(\theta), \,\, g_{0, i} = e^{a_{0, i} + i \phi_{0, i}}
    \end{aligned}
\end{equation}
Defining $V_{ij} = V_R + V_I$ we can reformulate and separate the real and imaginary parts as
\begin{align}
    \Delta[\text{Re}(V_{ij}] = V_R (a_i + a_j) - V_I (\phi_i - \phi_j)\\
    \Delta[\text{Im}(V_{ij}] = V_I (a_i + a_j) + V_R (\phi_i - \phi_j).
\end{align}
We can also stack all the gain values $\vec{\delta} = \begin{pmatrix}
    \vec{a} - \vec{a}_0\\
    \vec{\phi} - \vec{\phi}_0
\end{pmatrix} \in \mathbb{R}^{2 N}$ and factorize the equations into the following
\begin{equation}
    \vec{r} = J \vec{\delta} + \epsilon, \,\, \epsilon \sim \mathcal{N} (0, \Sigma_n), \,\, J \in \mathbb{R} ^{2M \times 2N},
    \label{eq: residual distribution}
\end{equation}
where again, $M$ and $N$ are the number of baselines and stations, respectively, $\epsilon$ is the stacked vector of all baseline thermal noises, and $J$ is the Jacobian matrix, which is defined as
\begin{equation}
    J = \begin{bmatrix}
        V_R S & - V_I D\\
        V_I S & V_R D
    \end{bmatrix} \in \mathbb{R}^{2M \times 2N}.
\end{equation}
With the following definitions for helper matrices $S$ and $D$.
\begin{equation}
    S_{mk} = \begin{cases}
        1, \,\,\, k = i(m) \text{ or } $k = j(m)$\\
        0, \,\,\, \text{otherwise}
    \end{cases}.
\end{equation}, i.e., it picks the sum $a_i + a_j$.

\begin{equation}
    D_{mk} = \begin{cases}
        +1, \,\,\, k = i(m)\\
        -1, \,\,\, k = j(m)\\
        0, \,\,\, \text{otherwise}
    \end{cases}.
\end{equation}
\subsection{Self-Calibration Via MAP Over Gains}
\label{supp:gain_marginalization}
The ultimate goal is to maximize the marginal likelihood 
\begin{equation}
    p(\vec{r}|\theta) = \int p(\vec{r}| \delta, \theta) p(\vec{\delta}) d\delta,
    \label{eq: marginal likelihood}
\end{equation}
which means to maximize the probability of the observed residuals given the neural network parameters $\theta$.
To do so, we first look at the elements of the integrand. The first element is the likelihood for complex visibilities given the gains $\vec{\delta}$ and the neural network parameters $\theta$
\begin{equation}
    p(\vec{r}|\vec{\delta}, \theta) \propto \exp(-\frac{1}{2} (\vec{r} - J \vec{\delta})^T \Sigma_n^{-1} (\vec{r} - J \vec{\delta})).
    \label{eq: residual| gains, nn params}
\end{equation}
For the other element in the integrand, we assume the prior distribution on gains to be another Gaussian of the form
\begin{equation}
        p(\vec{\delta}) \propto \exp(- \frac{1}{2} \vec{\delta}^T \Sigma_g^{-1} \vec{\delta}).
\end{equation}
Note, however, the form for these probabilities is not enough as the integral of the marginal likelihood is interactable. Instead of calculating the integral, however, we can calculate the MAP of $\vec{\delta}$ given $\vec{r}, \theta$, i.e., find the $\hat{\delta}$ that maximizes $p(\vec{\delta}| \vec{r}, \theta)$, and then expand the integral around $\hat{\delta}$. 
Bayes' for $\vec{r}$, $\delta$. and $\theta$ is
\begin{equation}
    p(\vec{\delta}| \vec{r}, \theta) = \frac{p(\vec{r}|\vec{\delta}, \theta) p(\vec{\delta})}{p(\vec{r}|\theta)}.
\end{equation}
And since the denominator is not dependent on $\vec{\delta}$, we have
\begin{equation}
    \begin{aligned}
        p(\vec{\delta}| \vec{r}, \theta) &\propto p(\vec{r}|\vec{\delta}, \theta) p(\vec{\delta})\\
        \rightarrow -\log p(\vec{\delta}|\vec{r}, \theta) &= \frac{1}{2} (\vec{r} - J \vec{\delta})^T \Sigma_n^{-1} (\vec{r} - J \vec{\delta}) \\
        &+ \frac{1}{2} \vec{\delta}^T \Sigma_g^{-1} \vec{\delta} + \text{const.}\\
        &\equiv \frac{1}{2} Q(\delta;\theta) + \text{const.}
    \end{aligned}
    \label{eq: MAP for delta}
\end{equation}
Minimizing the negative log-likelihood in Eq.~\eqref{eq: MAP for delta} yields
\begin{equation}
    \hat{\delta}(\theta) = H^{-1} J^T \Sigma_n^{-1} \vec{r}, \,\,\, H \equiv J^T \Sigma_n^{-1} J + \Sigma_g^{-1}.
    \label{eq: MAP solution for delta}
\end{equation}
Note that the $H$ matrix above is also the Hessian for $Q$ since $\nabla^2_\delta Q = H$. The first term in the Hessian is called the data curvature (or Fisher information matrix from the data) and the second term is the prior curvature.\\
It's clear now that Eq.~\eqref{eq: marginal likelihood} is equivalent to
\begin{equation}
    p(\vec{r}|\theta) \propto |\Sigma_g|^{-1/2} \int \exp(-\frac{1}{2} Q(\delta; \theta)) d\delta.
\end{equation}
We can show that, using Eq.~\eqref{eq: MAP for delta} and Eq.\eqref{eq: MAP solution for delta}, $Q(\delta; \theta)$ can be rewritten exactly as a function of $\hat{\delta}$:
\begin{equation}
    Q(\delta; \theta) = Q(\hat{\delta}; \theta) + (\delta - \hat{\delta})^T H (\delta - \hat{\delta})
\end{equation}
We therefore have
\begin{equation}
    \begin{aligned}
        p(r|\theta) &\propto |\Sigma_g|^{-1/2} \exp(-\frac{1}{2} Q(\hat{\delta}; \theta))\\
        & \int \exp(-\frac{1}{2} (\delta - \hat{\delta})^T H (\delta - \hat{\delta})) d\delta\\
        &\propto \exp(-\frac{1}{2} Q(\hat{\delta}; \theta)) |H|^{-\frac{1}{2}} |\Sigma_g|^{-1/2}    
    \end{aligned}
\end{equation}
And the negative likelihood is exactly what we use as the loss function for \neuralDMD:
\begin{equation}
    -\log p(\vec{r}|\theta) \propto Q(\hat{\delta}; \theta) + \log |H| + \log|\Sigma_g| \equiv \mathcal{L}.
    \label{eq: ndmd loss}
\end{equation}
With further algebraic manipulation (and noting that the priors are gains are constant) we can have the following equivalent forms for the above negative log-likelihood
\begin{equation}
    \begin{aligned}
        \mathcal{L} &= r^T \Sigma_n^{-1} r - \hat{\delta}^T H \hat{\delta} + \log |H| \\
        &= r^T \Sigma_n^{-1} r - \hat{\delta}^T J^T \Sigma_n^{-1} r + \log |H|
    \end{aligned}
    \label{eq: ndmd loss alternative forms}
\end{equation}
In practice, we use this marginalized likelihood as the objective in an outer Gauss--Newton scheme over $\theta$: at each iteration, given the current model visibilities we construct $J$, solve for $\hat{\delta}(\theta)$ via Eq.~\eqref{eq: MAP solution for delta}, and evaluate $\mathcal{L}$ in Eq.~\eqref{eq: ndmd loss alternative forms}. This is equivalent to an inner self-calibration loop over gains nested inside Gauss--Newton updates of \neuralDMD parameters.

\subsection{Numerical Stabilization}
There is a numerical stability challenge in directly optimizing the loss of Eq.~\eqref{eq: ndmd loss alternative forms} when the entries of $J$ matrix vary a lot in magnitude. We first describe this problem in detail and propose the solution. 
For each column of $J^T W J$, where we define $W \equiv \Sigma_n^{-1}$, we have
\begin{equation}
    s_k = ||j_k||_W = \sqrt{j_k^T W j_k}.
\end{equation}
Define $S = \text{diag}(s1, \dots, s_N)$ and, if the weighted inner product is $\rangle x, y\langle_W = x^T W y$,  the correlation matrix
\begin{equation}
    \begin{aligned}
        R_{kl} = \frac{\langle j_k, j_l\rangle_W}{s_k s_l} &\rightarrow J^T W J = S R S,\\
        \text{diag}(R) &= 1,\,\, |R_{kl}| \leq 1.
    \end{aligned}
    \label{eq: correlation matrix}
\end{equation}
Therefore, the Hessian $H$ is
\begin{equation}
    \begin{aligned}
        H &= J^T W J + \Sigma_g^{-1} = S R S + \Sigma_g^{-1},\\
        \Sigma_g^{-1} &= \text{diag}(\frac{1}{\sigma_{a_1}^2}, \dots, \frac{1}{\sigma_{a_N}^2}, \frac{1}{\sigma_{\phi_1}^2}, \dots, \frac{1}{\sigma_{\phi_N}^2}).
    \end{aligned}
\end{equation}
The diagonal terms are thus $H_{kk} = s_k^2 + \frac{1}{\sigma_k^2}$, and the off-diagonals are $H_{kl} = s_k s_l R_{kl}$. What goes wrong is that since the column norms $\{s_k\}$ have a large dynamic range as they are proportional to the observed visibilities, the entries of H inherit that range. 

We perform column normalization to make $H$ numerically stable. Define the column scaling $S = \text{diag}(s_k)$ and the rescaled Jacobian $\tilde{J} = J S^{-1}$ leads to the following data curvature 
\begin{equation}
     \rightarrow \tilde{J}^T W \tilde{J} = S^{-T} (J^T W J) S^{-1} = R,
\end{equation}
which means that now, the data curvature is equivalent to the numerically stable correlation matrix as defined in Eq.~\eqref{eq: correlation matrix}. 

The rescaling of the Jacobian $\tilde{J} = J S^{-1}$ must now be propagated through all variables for consistency. Starting from Eq.~\eqref{eq: residual distribution} (where we drop the arrow for expediency)
\begin{equation}
    r = J \, S^{-1} \tilde{\delta} + \epsilon.
    \label{eq: tilde residual distribution}
\end{equation}
In order for both equations to be equivalent, we must have $\tilde{\delta} = S \, \delta$. This new definition does not change Eq.~\eqref{eq: residual| gains, nn params}; however, it does change the gain priors
\begin{equation}
    p(\tilde{\delta}) \propto \exp{(-\frac{1}{2} \tilde{\delta}^T S^{-T} \Sigma_g^{-1} S^{-1} \tilde{\delta})}.
\end{equation}
Minimizing Eq.~\eqref{eq: MAP for delta} now leads to the solution
\begin{equation}
    \begin{aligned}
        \hat{\tilde{\delta}} &= S H^{-1} (J^T \Sigma_n^{-1} r + \Sigma_g^{-1} m) = S \, \hat{\delta},\\ 
        \tilde{H} &\equiv \tilde{J}^T \Sigma_n^{-1} \tilde{J} + S^{-T} \Sigma_g^{-1} S^{-1} = S^{-T} H \, S^{-1}.
    \end{aligned}
\end{equation}
Note that using the scaled variables (with tildes) as opposed to the original variables does not change the loss of Eq.\eqref{eq: ndmd loss alternative forms}.
% {
%     \small
%     \bibliographystyle{ieeenat_fullname}
%     \bibliography{main}
% }
\bibliographystyle{splncs04}
\bibliography{main}
\end{document}